\documentclass[a4paper,fleqn]{cas-sc}

\usepackage[numbers]{natbib}
\usepackage{hyperref}
\hypersetup{
  hidelinks,
  colorlinks=true,
  allcolors=black
}
\usepackage{mathtools}
\usepackage{booktabs}
\usepackage{multirow}
\usepackage{color}
\usepackage{blkarray}
\usepackage{graphicx}
\usepackage{subfig}
\graphicspath{{./}}
\usepackage{longtable}

\newdefinition{definition}{Definition}
\newcommand{\aka}{\textit{a}.\textit{k}.\textit{a}. }
\newcommand{\ie}{\textit{i}.\textit{e}. }
\newcommand{\eg}{\textit{e}.\textit{g}. }
\def\highlightcolor{black}
\newcommand{\marktext}[1]{\textcolor{\highlightcolor}{#1}}

\begin{document}

\title[mode=title]{\marktext{An original model for multi-target learning of logical rules for knowledge graph reasoning}}
\shorttitle{An original model for multi-target learning of logical rules for knowledge graph reasoning}

\author{Yuliang Wei}[style=chinese]
\ead{wei.yl@hit.edu.cn}
\author{Haotian Li}[style=chinese]
\ead{lcyxlihaotian@126.com}
\author{Guodong Xin}[style=chinese]
\ead{gdxin@hit.edu.cn}
\author{Yao Wang}[style=chinese]
\ead{wangyao_19941@hotmail.com}
\author{Bailing Wang}[style=chinese]
\ead{wbl@hit.edu.cn}
\cormark[1]
\cortext[cor1]{Corresponding author.}

\address{School of Computer Science and Technology, Harbin Institute of Technology at Weihai, China}

\begin{abstract}
Large-scale knowledge graphs provide structured representations of human knowledge.
However, as it is impossible to collect all knowledge, knowledge graphs are usually incomplete.
Reasoning based on existing facts paves a way to discover missing facts.
In this paper, we study the problem of learning logical rules for reasoning on knowledge graphs
for completing missing factual triplets. Learning logical rules equips a model with strong
interpretability as well as the ability to generalize to similar tasks. We propose a model able
to fully use training data which also considers multi-target scenarios.
In addition, considering the deficiency in evaluating the performance of models and the quality of
mined rules, we further propose two novel indicators to help with the problem.
Experimental results empirically demonstrate that our model outperforms state-of-the-art methods
on five benchmark datasets. The results also prove the effectiveness of the indicators.
\end{abstract}


\maketitle

\section{Introduction}
Knowledge storage, representation and its causal relationship between each other, inspired by
human problem solving, is to help intelligent systems understand human knowledge and gain the
ability to deal with complicated tasks \cite{newell1959report, mycin1976computer}.
Knowledge graphs (KGs), as a form of structured human knowledge, are collections of real-world factual triplets,
where each triplet $(s, p, o)$ denotes a predicate (\aka relation) between the subject $s$ and
the object $o$. Subjects and objects are usually called entities in KGs,
\textit{e.g.}, the fact that \emph{Beijing} is the \emph{capital} of \emph{China}
can be represented by (\emph{Beijing}, \emph{capitalOf}, \emph{China}). Knowledge graphs
are now widely used in a variety of applications such as recommender systems \cite{xian2019reinforcement, wang2019kgat}
and question answering \cite{lin2019kagnet, ding2019cognitive}. Recently knowledge graphs
have drawn growing interests in both academia and industry communities
\cite{dong2014knowledge, nickel2015review, sun2019rotate}.

\begin{figure}[pos=h, width=12cm, align=\centering]
  \centering
    \includegraphics[scale=.4]{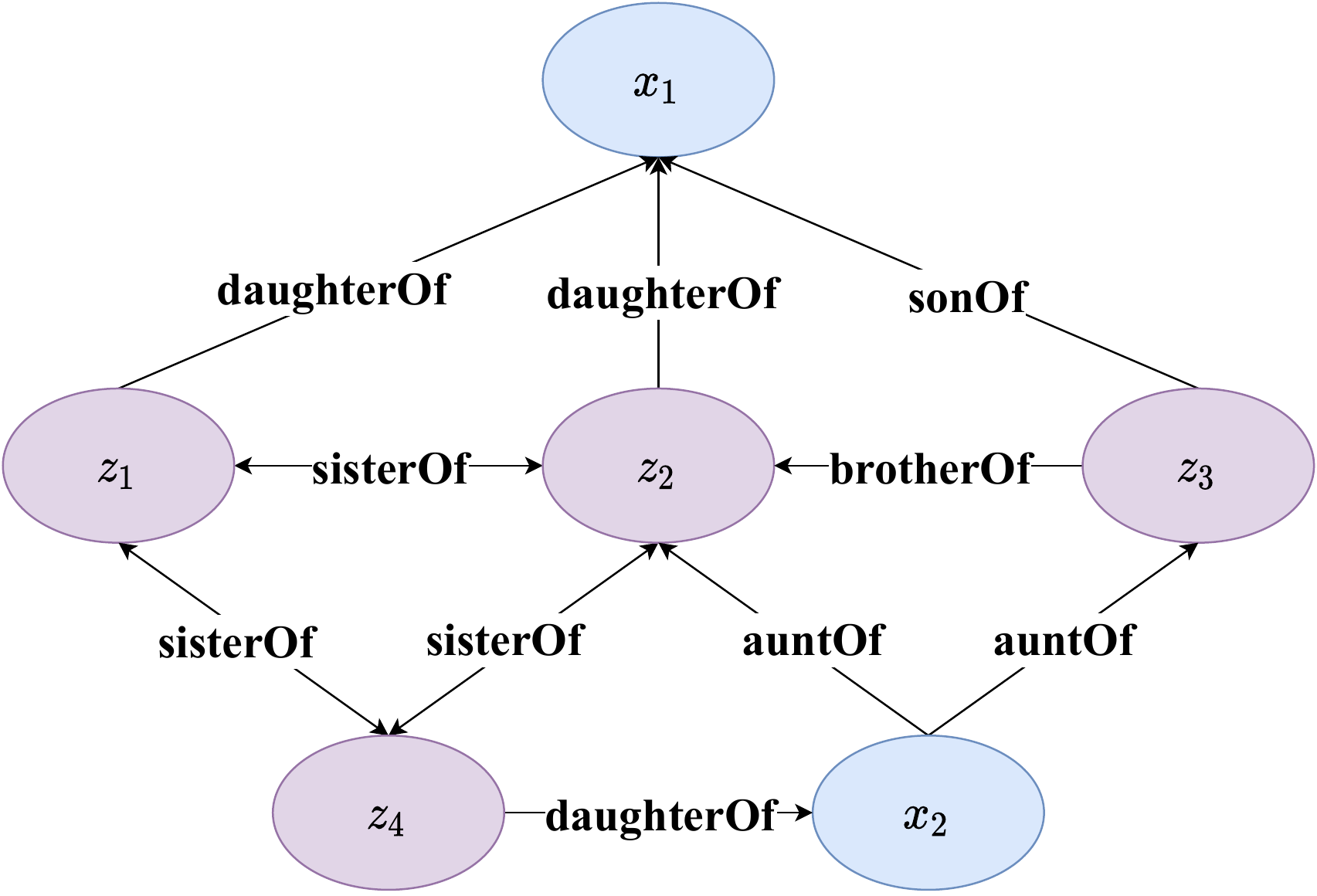}
  \caption{An example of knowledge graph reasoning in multi-target scenario.}
  \label{fig:multi-target}
\end{figure}

However, because of the nature of rapid iteration and incompleteness of data, there are usually missing
facts in existing KGs. For example, we now have the fact(s) that Thiago Messi, Mateo Messi and
Ciro Messi are sons of Leo Messi, but in our KG there might be missing information about the relationship
between the three brothers. A typical task is link prediction which is supposed to complete the
relation between two entities by reasoning on given facts. This paper studies learning
first-order logical rules for knowledge graph reasoning (KGR). As illustrated in Fig. \ref{fig:multi-target},
there is a rule in form of logic programming as \texttt{sisterOf}($X$, $Z$) $\wedge$
\texttt{sonOf}($Z$, $Y$) $\Rightarrow$ \texttt{daughterOf}($X$, $Y$),
meaning that if $X$ is a sister of $Z$ and $Y$ has a son $Z$,
then we can infer that $X$ is the daughter of $Y$. Such logical rules gain strong interpretability
\cite{qu2019probabilistic, zhang2020efficient} and can be applied for reasoning new facts and
generalized to previously unseen domains and data without retraining the model \cite{teru2020inductive}.
The same might not be true for embedding methods like TransE \cite{bordes2013translating}.

Mining collections of relational rules is a subtask of \emph{statistical relational learning}
\cite{koller2007introduction}, and when the procedure involves learning new logical rules,
it is often called \emph{inductive logic programming} \cite{muggleton1994inductive}.
Traditional methods such as Path Ranking \cite{lao2010relational}
and Markov Logic Networks \cite{richardson2006markov} failed to learn the structure
(\ie logical rules in discrete space) and the parameters (\ie continuous confidence associated
with each rule) simultaneously. The Neural LP method \cite{yang2017differentiable},
a fully end-to-end differentiable neural system, first combines learning rule structures
as well as appropriate scores. Unluckily, Neural LP and current Neural LP-based methods
lack attempts in multi-target scenarios, where there may be multiple objects
connecting by the same relation with one subject. Meanwhile, to the best of our knowledge,
although there are metrics for evaluating models in knowledge graph completion tasks,
there is still an absence in assessing the quality of mined logical rules.

In this paper, with reference to previous research in graph theory, we firstly propose two
novel indicators \emph{saturation} and \emph{bifurcation} that help with the evaluation
in KGR tasks. Saturation helps to investigate the interpretability of learned rules,
while bifurcation serves as a supplement to \marktext{traditional metrics on reasoning performance}.
Then we explore Multi-target Probabilistic Logic Reasoning (MPLR):
an extension to Neural LP framework that allows for reasoning in multi-target cases.
\marktext{Our approach reformulates the equations and improves the representation of entity and
the method to optimize the model, which enable our model to learn over more facts in the KG.}
We apply the indicators to several knowledge graph benchmarks for better understanding of 
their data structure. Further, we evaluate our model on these datasets and experimentally show that
our model outperforms state-of-the-art methods for knowledge graph reasoning.
In addition, MPLR is able to generate high-quality logical rules.

\marktext{Our work is related to previous efforts on the task of link prediction,
which can be categorized into two main streams: knowledge graph embedding and logical rule learning.}

\textbf{Knowledge graph embedding}. Preliminary research on knowledge graph completion focused on
learning low-dimensional embedding for link prediction, and we term those methods as embedding-based methods.
Representative methods, including TransE \cite{bordes2013translating}, TransR \cite{lin2015learning},
ComplEx \cite{trouillon2016complex}, etc., infer facts by projecting entities and relations
onto a semantic space and perform algebraic operations on that space.
Specifically, TransE \cite{bordes2013translating} presents factual triplets in $d\mbox{-}$dimensional
representation space, $s, p, o \in \mathbb{R}^d$ and makes embeddings follow the
translational principle $s + p \approx o$. TransR \cite{lin2015learning} tackles the problem of
insufficient representation ability of single space, and utilizes separate spaces for
entities and relations. ComplEx \cite{trouillon2016complex} is the first to introduce complex vector
space which can capture both symmetric and antisymmetric relations. In this space, $s, p, o \in \mathbb{C}^d$,
e.g., $s$ can be denoted as $s = \texttt{Re}(s) + i\texttt{Im}(s)$ where \texttt{Re}($s$) and
\texttt{Im}($s$) are real and imaginary parts of $s$ respectively.
Unfortunately, \marktext{one of the main
difficulties faced by embedding-based methods is the sparsity problem where their capability of
encoding sparse entities is far from satisfactory \cite{zhang2019iteratively}.
We also notice that recently there are methods trying to leverage logical rules into knowledge graph
embedding \cite{zhang2019iteratively, wangLogicRulesPowered2019}, where they explore new triplets
from the existing ones in the KG using pre-defined logical rules to deal with the sparsity problem.
}
However, these sort of methods are still implemented in a black-box way, which is uninterpretable to human.

\textbf{Relation path reasoning}. Learning relational rules has been previously studied in
the field of \emph{inductive logic programming} (ILP) \cite{muggleton1994inductive}. These methods
often learn a probability as a confidence score for each rule between query entities and answer entities.
Among these studies, Path-Ranking Algorithm (PRA) \cite{lao2010relational} enumerates relational
paths under a combination of path constraints and perform maximum-likelihood classification.
Markov Logic Networks \cite{richardson2006markov} and Probabilistic Personalized Page Rank (ProPPR)
\cite{wang2013programming} equip logical rules with probability, so that one can leverage path information
over the graph structure. Although ILP takes advantage of the interpretability of mined rules,
these methods typically require both positive and negative examples and suffer from
a potentially large version space, which is a critical shortage since most modern
KGs are huge and contain only positive instances.

\textbf{Neural logic programming}. Extending the idea by simultaneously learning logical rules and the
weights in a gradient-based way, Neural LP \cite{yang2017differentiable} is the first
end-to-end differentiable approach to combine continuous parameters and discrete structure of rules.
Some recent methods \cite{sadeghian2019drum, yang2019learn, wang2019differentiable} have improved the work
done by Neural LP \cite{yang2017differentiable} in different manners. DRUM \cite{sadeghian2019drum}
introduces tensor approximation for optimization and Neural-Num-LP \cite{wang2019differentiable}
addresses the limitation in mining numerical features like \emph{age} and \emph{weight}.
However, the existing Neural LP-based methods needs a large proportion of triplets
in preparation for constructing the graph structure, which can not make full use of the training data.
Moreover, these models \marktext{fail} in the situation of multi-target inference in contrast to our work.

\textbf{Related research in graph theory}. Although ILP shortens the gap between
reasoning on KGs and interpretability, there is still a lack of a way to
indicate the quality of learned rules. Inspired by accumulating studies \marktext{such as} k-saturated graphs
\cite{hajnal1965theorem} and minimum saturated graphs \cite{faudree2011survey}, we propose
\emph{saturation} concept as a complement in measuring the quality of rules.
Besides, we also define \emph{bifurcation} indicator to help with current metrics for evaluating
reasoning models from the perspective of graph structure.

\marktext{The main contributions of this work are summarized as follows:}
\begin{itemize}
  \item We propose two novel indicators as our extra performance metrics to evaluate models in experiment,
        one for reasoning accuracy on the task of knowledge graph completion, and the other for
        the feasibility of mined rules by our model.
  \item To tackle the problem of multi-target reasoning, which many existing methods fail to
        address, we develop a model based on logical rule learning, called MPLR. The proposed method
        represents knowledge in a multi-target form and develops a corresponding formulation to
        eliminate the side effects brought by learning over multiple factual triplets simultaneously.
  \item Extensive experiments on five benchmark datasets with traditional metrics as well as our
        proposed indictors prove that our MPLR model outperforms baseline models, and more importantly,
        is capable of mining meaningful rules from knowledge graphs.
\end{itemize}

\marktext{
The remainder of this paper is organized as follows. First, in Section \ref{sec:preliminaries},
we briefly review the basic concepts of knowledge graphs and propose the novel indicators by
giving definitions. Then, Section \ref{sec:method} introduces our proposed MPLR model for learning
logical rules on knowledge graphs in multi-target scenarios. Next, we conduct a series of comparative
experiments and the experimental results are reported in Section \ref{sec:experiment}. Finally,
we conclude our work together with our future direction.
}

\section{Preliminaries and two novel indicators}\label{sec:preliminaries}
\marktext{
In this section, we first introduce preliminary concepts and definitions of knowledge graphs
and knowledge graph reasoning. Then, two novel indicators for evaluating models and logical rules are
proposed in the rest of the section.
}

\subsection{Knowledge graph reasoning}
\textbf{Knowledge graph} can be modeled as a collection of
factual triplets $\mathcal{G} = \{ (s, p, o)\ |\ s, o \in \mathcal{E}, p \in \mathcal{P} \}$,
with $\mathcal{E}, \mathcal{P}$ representing the set of entities and predicates
(\aka binary relations) respectively in the knowledge graph, and $tri = (s, p, o)$ the triplet
$(subject, \texttt{predicate}, object)$ in form of $s \mathop{\rightarrow}\limits^p o$.
The subgraph relating to a particular predicate $\texttt{p}_i$ is described as a subset of
$\mathcal{G}$ containing all triplets with $\texttt{p}_i$ being the predicate:
$\mathcal{G}(\texttt{p}_i) = \{ (s, p, o)\ |\ s, o \in \mathcal{E}, \texttt{p}_i \in \mathcal{P},
p = \texttt{p}_i\}$.

\begin{definition}[Directed Labeled Multigraph]
A \textbf{directed labeled multigraph} $G$ is a tuple $G = (V, E)$,
where $V$ denotes the set of vertices, and $E \subseteq V \times V$
is a multiset of directed, labeled vertex pairs (\ie edges) in the graph $G$.
\end{definition}

Because of its graph structure, a knowledge graph can be regarded as a directed labeled multigraph
\cite{stokman1988structuring}. In this paper, "graph" is used to refer to "directed labeled multigraph"
for the sake of simplicity. $G(\texttt{p}) = (V(\texttt{p}), E(\texttt{p}))$
is the corresponding graph structure of $\mathcal{G}(\texttt{p})$.
$m = |V|$ and $n = |E|$ stand for the \textbf{number of vertices}
and \textbf{number of edges} respectively for a graph $G$.
Particularly in a KG, $|\mathcal{E}| = m$ and the total number of
triplets $(s, p, o)$ equals the number of edges $|\mathcal{G}| = n$.

In a graph $G = (V, E)$, the \textbf{degree} of a vertex $v \in V$ is the number of
edges incident to it. When it comes to directed graphs, \textbf{in-degree} and
\textbf{out-degree} of a vertex $v$ is usually distinguished, which are defined as
\begin{align}
deg^{+}(v) & = |\{ (u, v)\ |\ \exists u \in V, (u, v) \in E \}| \\
deg^{-}(v) & = |\{ (v, u)\ |\ \exists u \in V, (v, u) \in E \}|
\end{align}

Furthermore in KGs, the \textbf{bw-degree(q)} and \textbf{fw-degree(q)} of a given vertex $v$
can be computed via the following equations
\begin{align}
deg^{+}_q(v) & = |\{ (u, v)\ |\ \exists u \in V, (u, q, v) \in \mathcal{G} \}| \\
deg^{-}_q(v) & = |\{ (v, u)\ |\ \exists u \in V, (v, q, u) \in \mathcal{G} \}|,
\end{align}

\marktext{
which are exactly the number of entities connected by relation $q$ starting from $v$
when performing backward and forward reasoning.
}

\marktext{
In this work, the task of knowledge graph reasoning is regarded as a 
\textbf{probabilistic logic reasoning} \cite{wang2013programming} one, which is to learn a \emph{confidence} score
$\alpha \in [0,1]$ for a first-order logical \emph{rule} of the form
}
\begin{equation}
p_1(x, z_1) \wedge \cdots \wedge p_l(z_{l-1}, y) \Rightarrow q(x, y)\ : \ \alpha,
\end{equation}
$\textbf{p(x, y)} \Rightarrow q(x, y)$ for short, with $p_1, \ldots p_{l}, q \in \mathcal{P}$,
$z_i \in \mathcal{E}$, where $\textbf{p} = \wedge_i p_i$, is called a \texttt{rule pattern}.
For example, the rule $\texttt{brotherOf}(x, z) \wedge \texttt{fatherOf}(z, y) \Rightarrow \texttt{uncleOf}(x, y)$
intuitively states that if $x$ is the brother of $z$ and $z$ is the father of $y$,
then we can conclude that $x$ is the uncle of $y$.
All rule patterns of length $l$ ($l \geq 2$) can be formally defined as a set of predicate tuples
$\mathcal{H}^l = \{ (p_1, p_2, \ldots, p_l)\ |\ p_i \in \mathcal{P}, 1 \leq i \leq l \} = \mathcal{P}^l$,
and the set of patterns no longer than $L$ is denoted as
$\mathbb{H}^L = \mathop{\cup}\limits_{l=2}^L{\mathcal{H}^l}$.
A \texttt{rule path} $p$ is an instance of pattern \textbf{p} via different sequences of
entities, which is denoted as $p \rhd \textbf{p}$, \textit{e.g.}, $\left(p_a(x, z_1), p_b(z_1, y)\right)$ and
$\left(p_a(x, z_2), p_b(z_2, y)\right)$ are different paths of the same pattern.

\textbf{Multi-target reasoning.} Traditionally, the logic reasoning problem is to
solve the problem of learning first-order logical Horn clauses from a KG \cite{sadeghian2019drum}.
However, in multi-target scenarios, there would be several tail entities $y_i$ satisfying
the predicate $q$, given only one head entity $x$ such that $(x, q, y_i) \in \mathcal{G}$.
In other words, $x$ may have more than one nephew (niece) $y_i$,
and we may finally infer that $x$ is the uncle of $y$'s by following various rule patterns.

Therefore, the way we regard the knowledge graph reasoning task
is different from that of Neural LP \cite{yang2017differentiable}.
The task here is considered to be composed of a \texttt{query} $q \in \mathcal{P}$,
an entity \texttt{head} $h$ that the query is about,
and a set of entities \texttt{tails} $T$ that are the answers to the query
such that $(h, q, t) \in \mathcal{G}, \forall t \in T$. Finally we want to find the
most possible relational pattern $h \mathop{\rightarrow}\limits^{p_1} \cdots
\mathop{\rightarrow}\limits^{p_l} t, \forall t \in T$ to reason out the predicate $q$
through inducing over the whole query $(q, h, T)$.
Thus, given maximum length $L$, we assign a single \emph{confidence} score (\ie probability)
to a set of rule paths $p$'s adhering to the same pattern \textbf{p} that connects $h$ and $t$ \footnotemark:
\begin{equation}\label{eq:multi-target}
\{ p_i(h, t) \Rightarrow q(h, t)\ |\ p_i \rhd \textbf{p}, \textbf{p} \in \mathbb{H}^L,
t \in T \} \ : \ \alpha
\end{equation}

During inference, given an entity $h$, the unified score of a tail $t \in T$
can be computed by adding up the confidence scores of all rule paths that infer
$q(h, t)$, and the model will produce a ranked list of entities where higher the
score implies higher the ranking.

\footnotetext{
  In Neural LP framework, they view \texttt{tail}
  as the question to \texttt{query}, and only one \texttt{head} the answer to the query.
  Then a confidence $\alpha_i$ is assigned to one particular path $p_i$.
}

\subsection{Novel indicators for reasoning performance}\label{subsec:indicators}
\marktext{
In the following, we propose two novel metrics, \emph{saturation} and \emph{bifurcation},
}
to help evaluate a model for knowledge graph reasoning. More specifically, we analyze the reasoning complexity
from the inherent attributes of the graph structure $G$ corresponding to a KG $\mathcal{G}$.

\begin{definition}[Macro Reasoning Saturation]
Given a query $q \in \mathcal{P}$ and the maximum length $L$ of a rule pattern $\textbf{p}_l \in \mathbb{H}^L$,
the \textbf{macro reasoning saturation} of $\textbf{p}_l$ in relation to predicate $q$, \ie $\gamma^{\textbf{p}_l}_q$,
is the percentage of triplets $(h_i, q, t_j)$ in subgraph $\mathcal{G}(q)$
such that $\textbf{p}_l(h_i, t_j) \Rightarrow q(h_i, t_j)$.
\end{definition}

We compute the macro reasoning saturation $\gamma^{\textbf{p}_l}_q$ using the following equation:
\begin{equation}\label{eq:gamma}
\gamma^{\textbf{p}_l}_q = \frac{|\{ (h, q, t)\ |\ (h, q, t) \in \mathcal{G}(q),
\textbf{p}_l(h, t) \Rightarrow q(h, t) \}|} {n^q},
\end{equation}
with $n^q = |\mathcal{G}(q)|$ being the number of edges (\ie the number of triplets) in $G(q)$.
We can reasonably say that the larger $\gamma^{\textbf{p}_l}_q$ grows, the more likely $\textbf{p}_l$
can be as a proper inference of the query $q$. When $\gamma^{\textbf{p}_l}_q$ equals 1, it means
we can reason out every factual triplets in $\mathcal{G}(q)$ through at least one rule path
following the pattern $\textbf{p}_l$.

\begin{definition}[Micro Reasoning Saturation]
Given the maximum length $L$ of a rule pattern, we define the
micro reasoning saturation of pattern $\textbf{p}_l \in \mathbb{H}^L$ as following.
Firstly, for a specific triplet $\texttt{tri} = (h, q, t) \in \mathcal{G}$,
\ie $\delta^{\textbf{p}_l}_{\texttt{tri}}$, is the percentage of the number of paths
$\textbf{p}_{l_i} \rhd \textbf{p}_l$ such that $\textbf{p}_l(h, t) \Rightarrow q(h, t)$
as to all paths from $h$ to $t$.
\end{definition}

The equation to compute $\delta^{\textbf{p}_l}_{\texttt{tri}}$ is
\begin{equation}
\delta^{\textbf{p}_l}_{\texttt{tri}} = \frac{
|\{ \textbf{p}_{l_i}\ |\ \textbf{p}_{l_i} \rhd \textbf{p}_l, (h, q, t) \in \mathcal{G},
\textbf{p}_l(h, t) \Rightarrow q(h, t) \}|}
{|\{ \textbf{p}_{k_j}\ |\ \textbf{p}_{k_j} \rhd \textbf{p}_{k}, (h, q, t) \in \mathcal{G},
\forall \textbf{p}_{k} \in \mathbb{H}^L, \textbf{p}_{k}(h, t) \Rightarrow q(h, t) \}|}
\end{equation}

Then, we average $\delta^{\textbf{p}_l}_{\texttt{tri}}$ on all triplets $(h, q, t) \in \mathcal{G}(q)$
and get the \textbf{micro reasoning saturation} of pattern $\textbf{p}_l \in \mathbb{H}^L$
for query $q$:
\begin{equation}\label{eq:delta}
\delta^{\textbf{p}_l}_q = \frac{1}{n^q}\sum_{\texttt{tri} \in \mathcal{G}(q)}
{\delta^{\textbf{p}_l}_{\texttt{tri}}}
\end{equation}

In Eqs. (\ref{eq:gamma}) and (\ref{eq:delta}),
$\gamma^{\textbf{p}_l}_q$ and $\delta^{\textbf{p}_l}_q$ assess how easy it is to
infer $q$ following the pattern $\textbf{p}_l$ respectively from a macro and a micro perspective.
The higher the two indicators are, the easier we are to gain the inference
that $\textbf{p}_l(h, t) \Rightarrow q(h, t)$.
In order to obtain an overall result, we define the comprehensive reasoning saturation
$\eta^{\textbf{p}_l}_q$ by combining the two indicators through multiplication.
\begin{equation}
\eta^{\textbf{p}_l}_q = \gamma^{\textbf{p}_l}_q \times \delta^{\textbf{p}_l}_q
\end{equation}

\marktext{
The other indicator, \emph{bifurcation}, is proposed as follows.
}

\begin{definition}[Bifurcation]
Given a query $q$, the $\lambda$ \textbf{forward bifurcation} is the proportion of head entity $h \in V(q)$ 
with $fw\mbox{-}degree(q) \geq \lambda$ within all head entities in $\mathcal{G}(q)$.
Likewise, the $\lambda$ \textbf{backward bifurcation} is defined on tail entities in $\mathcal{G}(q)$
with $bw\mbox{-}degree(q) \geq \lambda$.
\end{definition}

Bifurcation(s) can be computed on both forward and backward reasoning directions and are formulated as follows:
\begin{align}
& fw\mbox{-}bifur^q(\lambda) = \frac{|\{ h\ |\ h \in V(q), deg^{-}_q(h) \geq \lambda \}|}
{|\{ h\ |\ h \in V(q) \}|} \\
& bw\mbox{-}bifur^q(\lambda) = \frac{|\{ t\ |\ t \in V(q), deg^{+}_q(t) \geq \lambda \}|}
{|\{ t\ |\ t \in V(q) \}|}
\end{align}

$fw\mbox{-}bifur^q(\lambda)$ and $bw\mbox{-}bifur^q(\lambda)$ indicate the problem scale
when performing backward and forward reasoning in case that there are multiple targets.
As is shown in Fig. \ref{fig:multi-target}, for query $q = \texttt{daughterOf}$,
there are three head entities $z_1, z_2, z_4$ and two tail entities $x_1$ and $x_2$. Hence the
$\lambda=2$ backward bifurcation of query $q$ is $bw\mbox{-}bifur^q(2) = 1/2 = 0.5$
for there are two ($\geq 2$) daughters  of $x_1$'s but only one of $x_2$'s,
meaning that half of the fathers (mothers) have at least two daughters. Similarly,
$fw\mbox{-}bifur^q(2) = 0/2 = 0$ because no one in $z_1, z_3, z_4$ has more than one parent
in this KG.

\section{\marktext{Multi-target learning of logical rules for knowledge graph reasoning}}\label{sec:method}
\marktext{
Combining learning structures and parameters, Neural LP \cite{yang2017differentiable} is the first
differentiable learning system for knowledge graph reasoning that learns representation and logical rules
simultaneously. Our work follows Neural LP and extensive studies based on it to consider the problem
of multi-target reasoning.
}

\subsection{Neural LP for logic reasoning}\label{subsec:neurallp}
\marktext{
Since Neural LP originally borrows the idea of the work of TensorLog
}
\cite{cohen2016tensorlog, kathryn2018tensorlog}, we first introduce TensorLog
that connects inference using logical rules with sparse matrix multiplication.
In a KG involving a set of entities $\mathcal{E}$ and a set of predicates $\mathcal{P}$,
factual triplets with respect to predicate $p_k$ are restored in a binary matrix $\mathrm{M}_{p_k}$
$\in \{0, 1\}^{|\mathcal{E}| \times |\mathcal{E}|}$. $\mathrm{M}_{p_k}$, an adjacency matrix,
is called a TensorLog operator meaning that $(e_i, p_k, e_j)$ is in the KG if and only if
the $(i, j)$-th entry of $\mathrm{M}_{p_k}$ is 1. Let $\mathrm{v}_{e_i} \in \{0, 1\}^{|\mathcal{E}|}$
be the one-hot encoded vector of entity $e_i$.
Then $s^{\top} = \mathrm{v}_{e_i}^{\top} \mathrm{M}_{p_1} \mathrm{M}_{p_2} \mathrm{M}_{p_3}$
is the \emph{path features vector} \cite{yang2019learn}, where the j-th entry counts the number of
unique paths following the pattern $p_1, p_2, p_3$ from $e_i$ to $e_j$ \cite{guu2015traversing}.

For example, every KG entity $e \in \mathcal{E}$ in Fig. \ref{fig:multi-target} is encoded into a
$0 \mbox{-} 1$ vector of length $|\mathcal{E}| = 6$. For every predicate $p \in \mathcal{P}$ and
every pair of entities $e_i, e_j \in \mathcal{E}$, the TensorLog operator relevant to $p$
is define as a matrix $\mathrm{M}_p$ with its $(i, j)$-th element being 1
\marktext{(highlighted by \textcolor{red}{red} in matrices)} if $(e_i, p, e_j) \in \mathcal{G}$.
Considering the KG in Fig. \ref{fig:multi-target}, for the predicate $p = \texttt{daughterOf}$ we have

\begin{equation*}
  \mathrm{M}_p =
  \begin{blockarray}{ccccccc}
    x_1 & x_2 & z_1 & z_2 & z_3 & z_4 \\
    \begin{block}{[cccccc]c}
      0 & 0 & 0 & 0 & 0 & 0 & x_1 \\
      0 & 0 & 0 & 0 & 0 & 0 & x_2 \\
      \textcolor{red}{1} & 0 & 0 & 0 & 0 & 0 & z_1 \\
      \textcolor{red}{1} & 0 & 0 & 0 & 0 & 0 & z_2 \\
      0 & 0 & 0 & 0 & 0 & 0 & z_3 \\
      0 & \textcolor{red}{1} & 0 & 0 & 0 & 0 & z_4 \\
    \end{block}
  \end{blockarray}
\end{equation*}

The rule \texttt{sisterOf}(X, Z) $\wedge$ \texttt{sonOf}(Z, Y) $\Rightarrow$ \texttt{daughterOf}(X, Y)
can be simulated by performing the following sparse matrix multiplication:

\begin{equation*}
  \mathrm{M_{p'}} = \mathrm{M_{sisterOf}}\ \mathrm{M_{daughterOf}} =
  \begin{blockarray}{ccccccc}
    x_1 & x_2 & z_1 & z_2 & z_3 & z_4 \\
    \begin{block}{[cccccc]c}
      0 & 0 & 0 & 0 & 0 & 0 & x_1 \\
      0 & 0 & 0 & 0 & 0 & 0 & x_2 \\
      \textcolor{red}{1} & \textcolor{red}{1} & 0 & 0 & 0 & 0 & z_1 \\
      \textcolor{red}{1} & \textcolor{red}{1} & 0 & 0 & 0 & 0 & z_2 \\
      0 & 0 & 0 & 0 & 0 & 0 & z_3 \\
      \textcolor{red}{1} & \textcolor{red}{1} & 0 & 0 & 0 & 0 & z_4 \\
    \end{block}
  \end{blockarray}
\end{equation*}
  
By setting $\mathrm{v}_{z_1} = [0, 0, 1, 0, 0, 0]^{\top}$ as the one-hot vector of $z_1$ and
multiplying by $\mathrm{v}_{z_1}^{\top}$ on the left,
we obtain $s^{\top} = \mathrm{v}_{z_1}^{\top} \cdot \mathrm{M_{p'}}$.
The resultant $s^{\top}$ selects the row in $\mathrm{M_{p'}}$ identified by $z_1$. By operating
right-hand side multiplication with $\mathrm{v}_{x_1}$, we get the number of unique paths
following the pattern $\texttt{sisterOf} \wedge \texttt{sonOf}$
from $z_1$ to $x_1$: $s^{\top} \cdot \mathrm{v}_{x_1} = 1$.

\textbf{Neural LP} \cite{yang2017differentiable} inherits the idea of TensorLog. Given a query
$q(h, t)$, after $L$ steps of reasoning, the score of the query induced through rule pattern
$\textbf{p}_s$ of length $L$ is computed as
\begin{equation}\label{eq:score}
\text{score}(t\ |\ q, h, \textbf{p}_s) =
\mathrm{v}^T_h \prod_{l=1}^L \mathrm{M}^l \cdot \mathrm{v}_t,
\end{equation}
where $\mathrm{M}^l$ is the adjacency matrix of the predicate used at $l$-th hop.

The operators above are used to learn for query $q$ by calculating the weighted sum
of all possible patterns:
\begin{equation}\label{eq:weightedSum}
\sum_s{\alpha_s \prod_{k \in \beta_s}{\mathrm{M}_{p_k}}},
\end{equation}
where $s$ indexes over all potential patterns with maximum length of $L$,
$\alpha_s$ is the confidence score associated with the rule $\textbf{p}_s$
and $\beta_s$ is the ordered list of predicates appearing in $\textbf{p}_s$.

To summarize, we update the score function in Eq. (\ref{eq:score})
by finding an appropriate $\alpha$ in
\begin{equation}\label{eq:varphi}
\varphi(t\ |\ q, h) =
\mathrm{v}^T_h \sum_s{\alpha_s \cdot \left(\prod_{k \in \beta_s}{\mathrm{M}_{p_k}} \cdot \mathrm{v}_t \right)},
\end{equation}
and the optimization objective is
\begin{equation}\label{eq:objective}
\max_{\alpha_s}{\sum_{(h, q, t) \in \mathcal{G}} {\varphi(t\ |\ q, h)}},
\end{equation}
where $\alpha_s$ is to be learned.

\begin{figure}[t]
  \centering
    \subfloat[]{
      \includegraphics[width=0.45\linewidth]{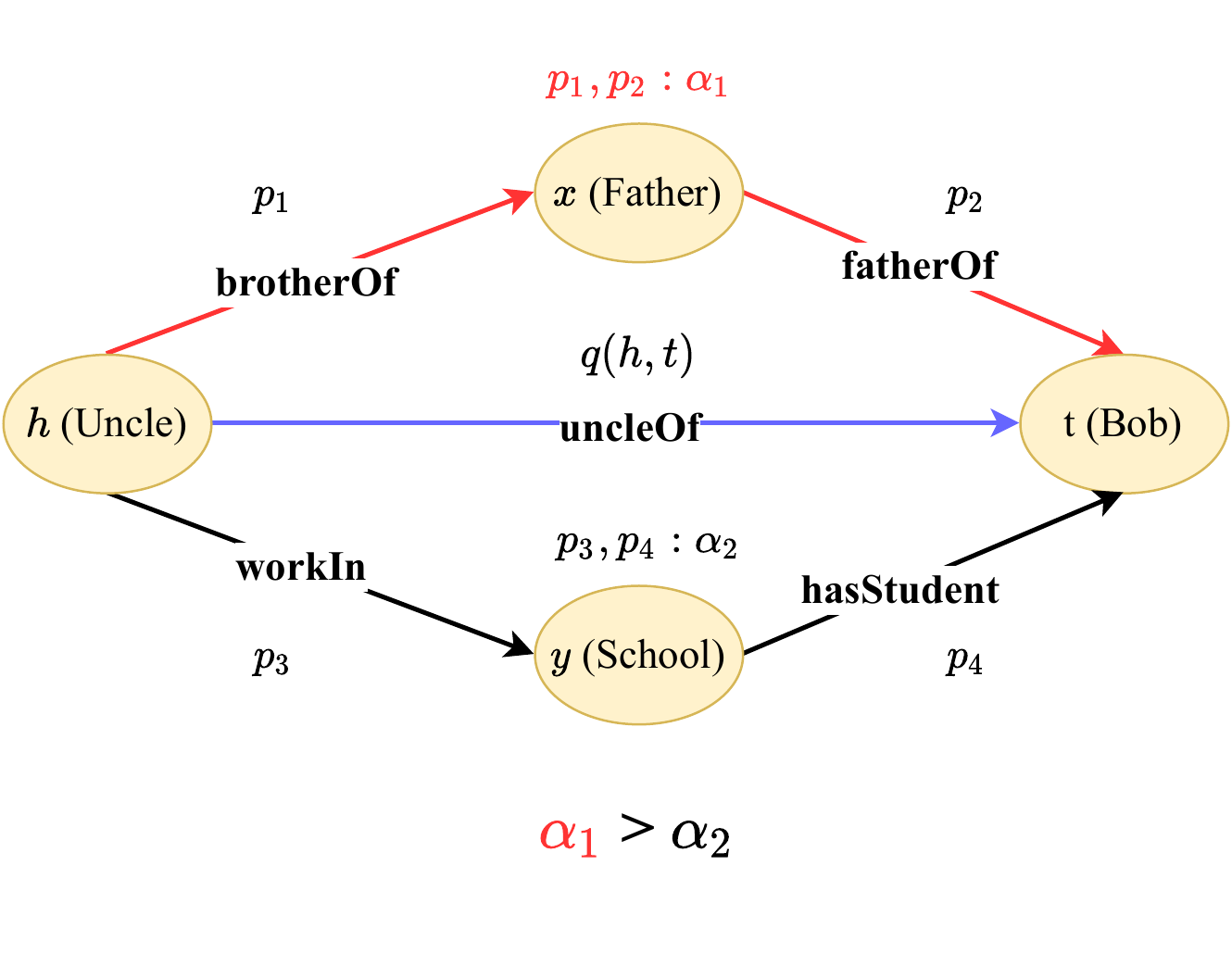}
      \label{fig:scoreFuncPhis(a)}
    }
    \hfill
    \subfloat[]{
      \includegraphics[width=0.45\linewidth]{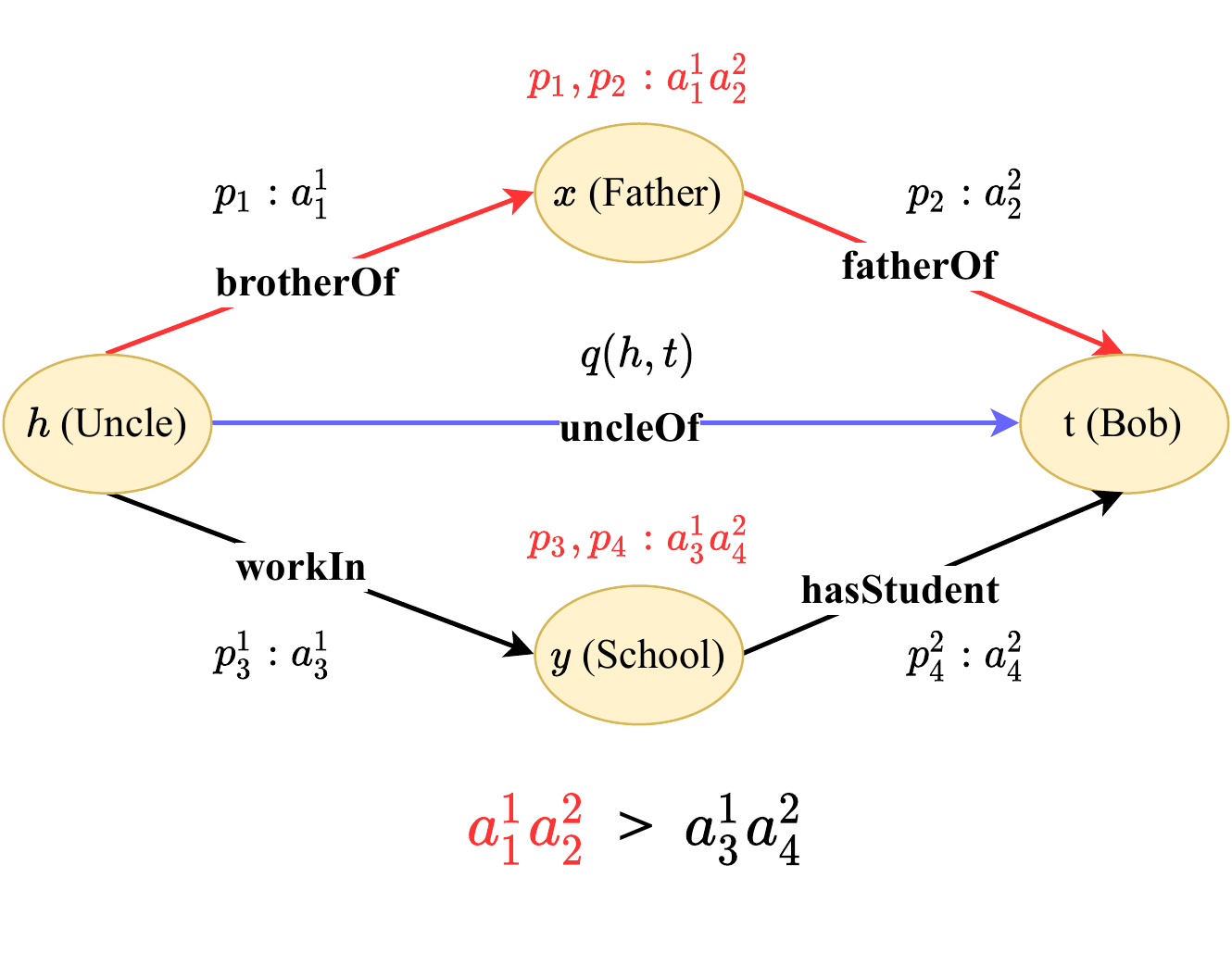}
      \label{fig:scoreFuncPhis(b)}
    }
  \caption{A KG example to illustrate the parameterization difference between
    Eqs (\ref{eq:varphi}) and (\ref{eq:phi}).
    (a) Assigning a confidence score to each rule. (b) Distributing weights
    into predicates at different hops.}
  \label{fig:scoreFuncPhis}
\end{figure}

Whereas the searching space of learnable parameters is exponentially large,
\ie $O(|\mathcal{P}|^L)$, direct optimization of Eq. (\ref{eq:objective})
may fall in the dilemma of over-parameterization. Besides,
it is difficult to apply gradient-based optimization. This is because each variable $\alpha_s$ is
bound with a specific rule pattern, and it is obviously a discrete work to enumerate rules.
To overcome these defects, the parameter of rule $\textbf{p}_s$ can be reformulated by distributing
the confidence to its containing predicate at each hop, resulting in a differentiable score function:
\begin{equation}\label{eq:phi}
\phi_L(t\ |\ q, h) = \left(\mathrm{v}^T_h \prod_{l=1}^{L}
{\sum_{k=0}^{|\mathcal{P}|}{a_k^l \mathrm{M}_{p_k}}} \right) \cdot \mathrm{v}_t,
\end{equation}
where $L$ is a hyperparameter denoting the maximum length of patterns and $|\mathcal{P}|$ is the
number of predicates in KG. $\mathrm{M}_{p_0}$ is an identity matrix $I$ that enables the model
to include all possible rule patterns of length $L$ or smaller \cite{sadeghian2019drum}.
The key difference of parameterization between Eq. (\ref{eq:varphi})
and Eq. (\ref{eq:phi}) is illustrated in Fig. \ref{fig:scoreFuncPhis}. Fig. \ref{fig:scoreFuncPhis(a)}
shows that the rule \texttt{brotherOf} $\wedge$ \texttt{fatherOf} gains more plausibility against
\texttt{workIn} $\wedge$ \texttt{hasStudent} facing the query \texttt{uncleOf}, so $\alpha_1 > \alpha_2$.
In the latter, the score of $\texttt{brotherOf} \wedge \texttt{fatherOf}$ is obtained by multiplying
the weight of \texttt{brotherOf} at first hop $a_1^1$ and \texttt{fatherOf} at second hop $a_2^2$.

To perform training and prediction over the Neural LP framework, we should first construct a KG from
a large subset of all triplets. Then we remove the edge $(h, t)$ from the graph
when facing the query $(h, q, t)$, so that the score of $t$ can get rid of the influence imposed
by passing from head entity $h$ directly through edge $(h, t)$ for the correctness of reasoning.

\subsection{\marktext{Our} MPLR model}\label{subsec:MPLR}

In this section, we propose our MPLR model as an improvement to Neural LP \cite{yang2017differentiable}.

As aforementioned in Section \ref{subsec:neurallp}, all edges starting from the head entity $h$
to $t_i$ should be removed from the graph in a multi-target query, thus more edges in a batch of
queries will be removed, which would break the graph structure to a considerable extent.
For example, in Fig. \ref{fig:multi-target}, suppose the query is $(z_1, \texttt{sisterOf}, \{z_2, z_4\})$,
so that two edges $(z_1, z_2)$ and $(z_1, z_4)$ will be missing when we train the model,
which renders difficulty to infer the rule \texttt{sisterOf}($z_1$, $z_2$) $\wedge$
\texttt{sisterOf}($z_2$, $z_4$) $\Rightarrow$ \texttt{daughterOf}($z_1$, $z_4$).
Therefore, we update Eq. (\ref{eq:phi}) to address the limitation of Neural LP in multi-target scenario,
where the bonus on the score of $t_i$ from edge $(h, t_i)$ is avoided without removing the edge.
For each sub-query
$(h, q, t), \forall t \in T$ we have
\begin{gather}
\mathrm{u}_0 = \mathrm{v}_h,\ \mathrm{\varepsilon}_1 = a_q^1 \cdot \mathrm{v}_t \\
\mathrm{u}_l^{\top} = \mathrm{u}_{l-1}^{\top}
\sum\limits_{k=0}^{|\mathcal{P}|}{a_k^l \mathrm{M}_{p_k}},\ l = 1, 2, \ldots, L \\
\mathrm{\varepsilon}_l^{\top} = \mathrm{\varepsilon}_{l-1}^{\top}
(\sum\limits_{k=0}^{|\mathcal{P}|}{a_k^l \mathrm{M}_{p_k}}) +
a_q^l \mathrm{u}_{l-1}^{\top} \mathrm{M}_t,\ l = 2, 3, \ldots, L \\
\varPhi_L(t\ |\ q, h) = (\mathrm{u}_L^{\top} - \mathrm{\varepsilon}_L^{\top} \mathrm{M}_t)
\cdot \mathrm{v}_t, \label{eq:varPhi}
\end{gather}
where $a_q^l$ is the attention score of predicate $q$ at $l$-th hop,
and $\mathrm{M}_t \in \{0,1\}^{|\mathcal{E}|}$ is the matrix
with only its $(t, t)$-th element being 1, otherwise 0.
$\mathrm{\varepsilon}_1^{\top}$ is actually vector $\mathrm{u}_1^{\top}$
with all elements reduced to 0 except its $t$-th value.
Also, vector $\mathrm{u}^{\top}_{l-1} \mathrm{M}_t$ keeps only the $t$-th value of $\mathrm{u}^{\top}_{l-1}$.

Eq. (\ref{eq:varPhi}) eliminates redundant gain (\ie $\mathrm{\varepsilon}_L^{\top}\mathrm{M}_t$)
of $t$-th value in vector $\mathrm{u}_L^{\top}$
passed from $h$ to $t$ directly through edge $(h, t)$, but retains the approach to affect other
nodes except $t$ through this edge. That is to say, in query ($z_1$, \texttt{sisterOf}, $\{z_2, z_4\}$)
the score of entity $z_2$ should not involve that from edge $(z_1, z_2)$, but the score of $z_4$
can be increased by path \texttt{sisterOf}($z_1$, $z_2$) $\wedge$ \texttt{sisterOf}($z_2$, $z_4$).

In addition, considering there might be multiple tail entities in relation to the given head entity in a query,
and as shown in Eq. (\ref{eq:multi-target}), only one score should be allocated to a set of rule paths.
We modify the representation of the \texttt{tail} vector to a multi-hot one.
Given a query $q$, a head entity $e_i$ and a set of tails $T$,
the target vector $\mathrm{v}_T \in \{ 0, 1 \}^{|\mathcal{E}|}$ is also a
$0 \mbox{-} 1$ vector, but with it $j$-th entry being 1 for all $e_j \in T$.
For example, in the KG displayed in Fig. \ref{fig:multi-target}, given query \texttt{auntOf}
and head entity $x_2$, since $x_2$ is aunt of $z_2$ and $z_3$,
the target vector in this query is $\mathrm{v}_T = [0, 0, 0, 1, 1, 0]^{\top}$.

Finally, the confidence scores are learned over the bidirectional LSTM \cite{hochreiter1997long}
followed by the attention using Eqs (\ref{eq:lstm}) and (\ref{eq:attn}) for the temporal dependency
among several consecutive steps. The \texttt{input} in Eq. (\ref{eq:lstm})
is \texttt{query} embedding for $1 \leq i \leq L$.
\begin{align}
& \textbf{h}_i, \textbf{h}'_{L-i+1} =
\text{BiLSTM}(\textbf{h}_{i-1}, \textbf{h}'_{L-i}, \text{input})\label{eq:lstm} \\
& [a_{i, 1}, \ldots, a_{i, |\mathcal{P}|}] =
f_{\theta}\left( [\textbf{h}_i\ ||\ \textbf{h}'_{L-i}] \right), \label{eq:attn}
\end{align}
where $\textbf{h}$ and $\textbf{h}'$ are the hidden-states of the forward and backward path LSTMs,
and the subscripts denote their time step. $[a_{i, 1}, \ldots, a_{i, |\mathcal{P}|}]$ is the
attention vector obtained by performing a linear transformation over concatenated forward and
backward hidden states, followed by a softmax operator: $f_{\theta}(H) = \text{softmax}
(WH + b)$.

\begin{figure*}[t]
  \centering
  \includegraphics[width=\linewidth]{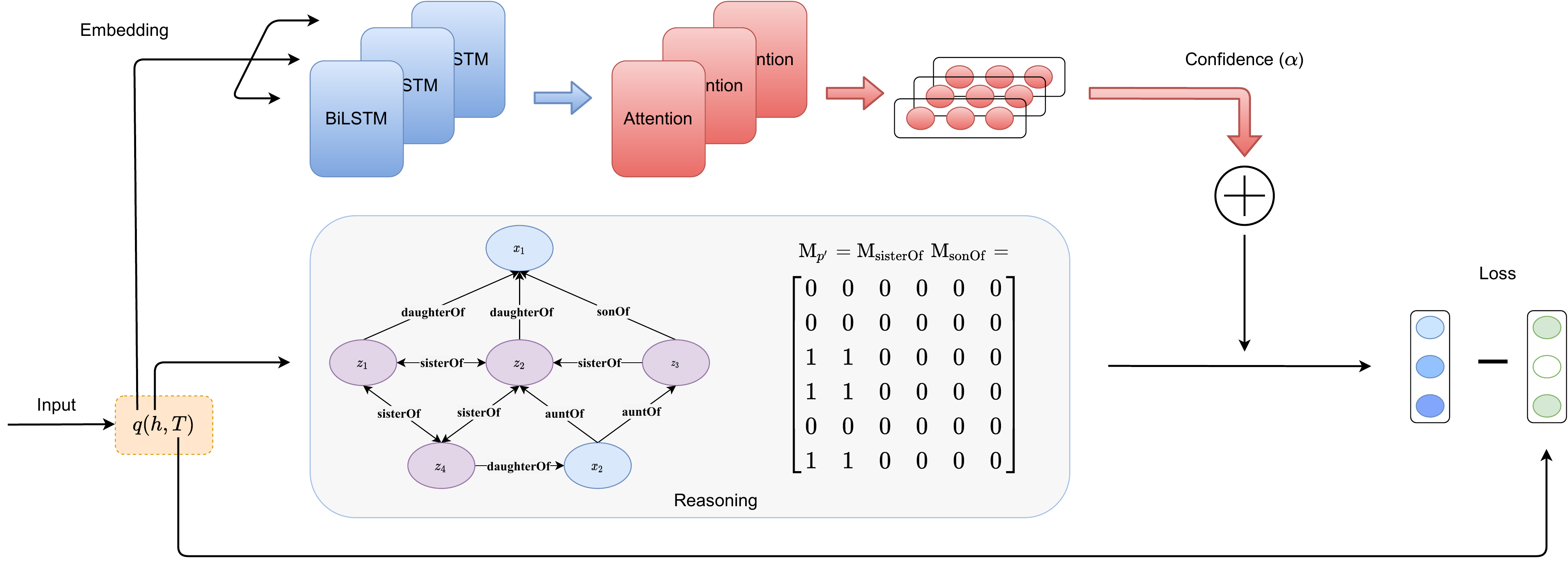}
  \caption{MPLR model overview with rank $R=3$.}
  \label{fig:model-overall}
\end{figure*}

\subsection{Optimization of the model}
\marktext{
As our work extends the fully differentiable framework, Neural LP \cite{yang2017differentiable},
a gradient-based algorithm ADAM \cite{kingma2014adam} is applied to optimize our model, which is
widely used in a large number of deep learning models. In this section, loss function and an optimization
method of tensor called lower-rank approximation are introduced respectively.
}

\textbf{Loss construction.} In general, we treat this task as a multi-label classification
to handle multiple outcomes.
For each query $q(h, T)$ in KG, we first split the objective function Eq. (\ref{eq:varPhi}) into
two parts: target vector $\mathrm{v}_T$ and prediction vector
\begin{equation}
\mathrm{s}^{\top} = \mathrm{u}_L^{\top} - \mathrm{\varepsilon}_L^{\top} \mathrm{M}_t,
\end{equation}
and then we construct the loss function for $\mathrm{v}_T$ and $\mathrm{s}^{\top}$
using the Bernoulli negative log-likelihood with logits:
\begin{equation*}
\ell_q(h, T) = -\sum_{i=1}^{|\mathcal{E}|} \left\{ \mathrm{v}_T[i] \cdot \log{\left( \sigma(\mathrm{s}[i]) \right)}
+ (1-\mathrm{v}_T[i]) \cdot \log{\left( \sigma(1-\mathrm{s}[i]) \right)} \right\},
\end{equation*}
where $i$ indexes elements in vector $\mathrm{v}_T$ and $\mathrm{s}$, and $\sigma(\cdot)$ is the
sigmoid function $\sigma(z) = \frac{1}{1+e^{(-z)}}$. To ensure numerical stability,
the above equation can be reformulated into the equivalent Eq. (\ref{eq:ell}) through log-sum-exp method.
\begin{equation}\label{eq:ell}
\ell_q(h, T) = \sum_{i=1}^{|\mathcal{E}|} \left\{ \max(\mathrm{s}[i], 0)
- \mathrm{v}_T[i] \cdot \mathrm{s}[i] + \log{(1+e^{-|\mathrm{s}[i]|})} \right\}
\end{equation}

\textbf{Low-rank approximation.} It can be shown that the final confidences obtained by expanding
$\varPhi_L$ are a rank one estimation of the \emph{confidence value tensor} \cite{sadeghian2019drum},
and a low-rank approximation is a popular method for tensor approximation. Hence we follow the work
of \cite{sadeghian2019drum} and rewrite Eq. (\ref{eq:varPhi}) using rank $R$ approximation,
as shown in Eq. (\ref{eq:objective'}).
\begin{equation}\label{eq:objective'}
\varPhi_L(t\ |\ q, h) = \sum_{r=1}^R{\left( \mathrm{u}_L^{\top} -
\mathrm{\varepsilon}_L^{\top} \mathrm{M}_t \right)} \cdot \mathrm{v}_t
\end{equation}

More concretely, we update Eqs. (\ref{eq:lstm}) and (\ref{eq:attn}),
as is shown in Eqs. (\ref{eq:lstm'}) and (\ref{eq:attn'}),
by deploying number of $R$ BiLSTMs of the same network structure,
each of which can extract features from various dimensions.
\begin{align}
& \textbf{h}_i^{(r)}, \textbf{h}_{L-i+1}^{'(r)} =
\text{BiLSTM}_r(\textbf{h}_{i-1}^{(r)}, \textbf{h}_{L-i}^{'(r)}, \text{input})\label{eq:lstm'} \\
& [a_{i, 1}^{(r)}, \ldots, a_{i, |\mathcal{P}|}^{(r)}] =
f_{\theta}\left( [\textbf{h}_i{'(r)}\ ||\ \textbf{h}_{L-i}^{'(r)}] \right), \label{eq:attn'}
\end{align}
where the superscripts of the hidden states identify their bidirectional LSTM.

An overview of the model is shown in Fig. \ref{fig:model-overall}.

\section{Experiment}\label{sec:experiment}
\subsection{Experiment setting}
We conduct experiments on a knowledge graph completion task and evaluate our model
in comparison with state-of-the-art baselines regarding the following aspects:
(1) traditional evaluation metrics (\eg Mean Reciprocal Rank);
(2) novel reasoning indicators proposed in Section \ref{subsec:indicators};
(3) interpretability, \textit{i.e.} reasoning plausibility.
After detailed explanations, the deficiency of the existing Neural LP-based models is also discussed.

\subsubsection{Datasets}\label{subsubsec:datasets}

We adopt five datasets for evaluation, which are described as follows:

\begin{itemize}
  \item \emph{FB15K-237} \cite{toutanova2015observed}, a more challenging version of
    FB15K \cite{bordes2013translating} based on Freebase \cite{bollacker2008freebase},
    a growing knowledge graph of general facts.
  \item \emph{WN18} \cite{dettmers2018convolutional}, a subset of knowledge graph
    WordNet \cite{miller1995wordnet, miller1998wordnet} constructed for a widely used dictionary.
  \item \emph{Medical Language System (UMLS)} \cite{kok2007statistical}, from biomedicine,
    where the entities are biomedical concepts (\eg \texttt{organism}, \texttt{virus}) and relations
    consist of \texttt{affects} and \texttt{analyzes}, etc.
  \item \emph{Kinship} \cite{kok2007statistical}, containing kinship relationships
    among members of a Central Australian native tribe.
  \item \emph{Family} \cite{kok2007statistical}, containing individuals from multiple families
    that are biologically related.
\end{itemize}

Statistics about each dataset are shown in Table \ref{tab:data-statistics}.
All datasets are divided into 3 files: \emph{train}, \emph{valid} and \emph{test}.
The \emph{train} file is composed of query examples $q(h, T)$.
\emph{valid} and \emph{test} files both contain queries $q(h, t)$,
in which the former is used for early stopping and the latter is for testing.
Unlike the case of learning embeddings, our method does not necessarily require the entities in
\emph{train}, \emph{valid} and \emph{test} to overlap.
As described in Section \ref{subsec:MPLR}, our model is capable of using all triplets
(serve as \emph{facts} file in Neural LP \cite{yang2017differentiable}) to construct KG,
including ones from \emph{train}, \emph{valid} and \emph{test}.

\subsubsection{Comparison of algorithms}\label{subsubsec:algorithms}
In experiment, the performance of our model is compared with that of the following algorithms:
\begin{itemize}
  \item Neural LP-based methods. Since our model is based on Neural LP \cite{yang2017differentiable},
    we choose Neural LP and a Neural LP-based method DRUM \cite{sadeghian2019drum}.
  \item Embedding-based methods. We choose several embedding-based algorithms, including
    TransE \cite{bordes2013translating}, DistMult \cite{yang2014embedding},
    TuckER \cite{balavzevic2019tucker}, RotatE \cite{sun2019rotate} and ConvE \cite{dettmers2018convolutional}.
  \item Other rule learning methods. We also consider a probabilistic model
    called RNNLogic\footnotemark \cite{qu2020rnnlogic}.
\end{itemize}

\footnotetext{There are four variants of RNNLogic, and we use RNNLogic without embedding for comparison.}

\subsubsection{Model configuration}
Our model is implemented using PyTorch \cite{paszke2019pytorch}.
We use the same hyperparameter suite during experiments on all datasets. The hidden state dimension
for BiLSTM(s) is 128. The query embedding has dimension 128 and is randomly initialized.
As for optimization algorithm, we use mini-batch ADAM \cite{kingma2014adam}
with the batch size 128 and the learning rate initially set to 0.001. We also observe that the whole
model tends to be more trainable if we normalize the vector $\mathrm{u}_l$ at final step to have
unit length.

\begin{table*}[pos=t, width=.75\linewidth]
  \caption{Statistics of datasets.}
  \label{tab:data-statistics}
  \centering
  \begin{tabular*}{\tblwidth}{ccccccc}
    \toprule
    Dataset & \# Relation & \# Entity & \# Triplets & \# Train & \# Validation & \# Test \\
    \midrule
    FB15K-237 & 237 & 14541 & 310116 & 272115 & 17535 & 20466 \\
    WN18 & 18 & 40943 & 151442 & 141442 & 5000 & 5000 \\
    Family & 12 & 3007 & 28356 & 23483 & 2038 & 2835 \\
    Kinship & 25 & 104 & 10686 & 8487 & 1099 & 1100 \\
    UMLS & 46 & 135 & 6529 & 5327 & 569 & 633 \\
    \bottomrule
  \end{tabular*}
\end{table*}

\subsection{Experiment on knowledge graph completion}

We conduct experiments on the knowledge graph completion task as described in \cite{bordes2013translating},
and compare the results with several state-of-the-art models. When training the model, the \texttt{query}
and \texttt{head} are part of some missing training triplets, and the goal is to complete the question
and find the most possible answers \texttt{tails}.
For example, if \texttt{daughterOf}($X$, $\{Y_1, Y_2, Y_3\}$) is missing from the knowledge graph\footnotemark,
the goal is to reason over the existing graph structure and retrieve $\{Y_1, Y_2, Y_3\}$ when
presented with query \texttt{daughterOf} and $X$.

\footnotetext{To be more accurate, our model simulates this situation that the edges
relating to the input query are removed, which is already explained in Section \ref{subsec:MPLR}.}

During evaluation, for each test triplet $(h, q, t)$, we build one query $(h, q, ?)$
with answer $t$ \footnotemark. Remarkably, we adopt the same \emph{valid} and \emph{test} data
with compared algorithms, and we manually remove the edge $(h, t)$ from KG for the correctness
of reasoning results. Additionally, when computing the actual rank of $t$,
the head entity $h$ is of no use in a query, so we manually remove it.
For each query, the score is computed for each entity, as well as the rank of the correct answer.
For the computed ranks from all queries, we report the Mean Reciprocal Rank (MRR) and Hit@$k$.
MRR averages the reciprocal rank of the answer entities and
Hit@$k$ computes the percentage of how many desired entities are ranked among top $k$.

\footnotetext{We notice that in Neural LP \cite{yang2017differentiable},
DRUM \cite{sadeghian2019drum}, etc., they add another reversed query $(?, q, t)$
with answer $h$ for each triplet. But we only use query $(h, q, ?)$ for fair comparison.}

\subsubsection{\marktext{Novel indicators on selected datasets}}

We calculate the numerical features of KG datasets using the indicators
proposed in Section \ref{subsec:indicators}, which helps to better comprehend
the reasoning task over these knowledge graphs. Above all, considering that learning collections of 
relational rules is a type of \emph{statistical relational learning} \cite{koller2007introduction},
these statistical properties provide a complement to currently popular evaluation metrics, such as
MRR and Hit@$k$.

\textbf{Saturation}. The \emph{macro}, \emph{micro} and \emph{comprehensive} saturations
measure the probability of a rule pattern occurring in a certain
relational subgraph $\mathcal{G}(p)$ from different angles. However, the computation can be exceedingly
costly due to the approximate complexity $\mathcal{O}(|\mathcal{P}| \cdot |\mathcal{P}|^L \cdot
|\mathcal{G}| \cdot \omega) = \mathcal{O}(|\mathcal{P}|^{L+1} \cdot |\mathcal{G}| \cdot \omega)$, where
$|\mathcal{P}|^L$ is the size of rule set $\mathcal{H}^L$, \textit{i.e.}, the total number of rules of length $L$,
and $\omega$ indicates how time-consuming to compute the number of unique paths following
pattern $\textbf{p}_l$ pointing from $h$ to $t$ given $\textbf{p}_l$ and $(h, q, t)$.
Thus, it is more preferable to randomly sample a subgraph of the existing KG first,
and then compute the saturations when encountering a large dataset.
We select some predicates and their relating rules with most popular saturations from the Family dataset
and show them in Table \ref{tab:saturation-family}. We also present the statistics about UMLS
in the appendix.

\begin{center}
\begin{longtable}{rclccc}
  \caption{
    Saturations of the Family dataset (without sampling).
    The rule length is fixed to 2. $\gamma^{\textbf{p}_l}_q$, $\delta^{\textbf{p}_l}_q$,
    $\eta^{\textbf{p}_l}_q$ are \emph{macro}, \emph{micro} and \emph{comprehensive} saturations.
    The results relating to a predicate are sorted by the comprehensive saturation in descending order.
  }
  \label{tab:saturation-family} \\
    \toprule
    Rule & $\Rightarrow$ & Predicate & $\gamma^{\textbf{p}_l}_q$ &
      $\delta^{\textbf{p}_l}_q$ & $\eta^{\textbf{p}_l}_q$ \\
    \midrule
    $X \xRightarrow{\texttt{motherOf}} Z \xRightarrow{\texttt{sonOf}} Y$ &
    $\Rightarrow$ & $X \xRightarrow{\texttt{wifeOf}} Y$ & .47 & .35 & .17 \\
    $X \xRightarrow{\texttt{motherOf}} Z \xRightarrow{\texttt{daughterOf}} Y$ &
    $\Rightarrow$ &  & .36 & .24 & .09 \\
    \midrule
    $X \xRightarrow{\texttt{fatherOf}} Z \xRightarrow{\texttt{sonOf}} Y$ &
    $\Rightarrow$ & $X \xRightarrow{\texttt{husbandOf}} Y$ & .47 & .35 & .17 \\
    $X \xRightarrow{\texttt{fatherOf}} Z \xRightarrow{\texttt{daughterOf}} Y$ &
    $\Rightarrow$ &  & .36 & .24 & .09 \\
    \midrule
    $X \xRightarrow{\texttt{wifeOf}} Z \xRightarrow{\texttt{fatherOf}} Y$ &
    $\Rightarrow$ & $X \xRightarrow{\texttt{motherOf}} Y$ & 1. & .34 & .34 \\
    $X \xRightarrow{\texttt{motherOf}} Z \xRightarrow{\texttt{brotherOf}} Y$ &
    $\Rightarrow$ &  & .70 & .27 & .19 \\
    $X \xRightarrow{\texttt{motherOf}} Z \xRightarrow{\texttt{sisterOf}} Y$ &
    $\Rightarrow$ &  & .62 & .22 & .14 \\
    \midrule
    $X \xRightarrow{\texttt{sisterOf}} Z \xRightarrow{\texttt{sonOf}} Y$ &
    $\Rightarrow$ & $X \xRightarrow{\texttt{daughterOf}} Y$ & .68 & .25 & .17 \\
    $X \xRightarrow{\texttt{sisterOf}} Z \xRightarrow{\texttt{daughterOf}} Y$ &
    $\Rightarrow$ &  & .61 & .20 & .12 \\
    $X \xRightarrow{\texttt{daughterOf}} Z \xRightarrow{\texttt{husbandOf}} Y$ &
    $\Rightarrow$ &  & .46 & .15 & .07 \\
    $X \xRightarrow{\texttt{daughterOf}} Z \xRightarrow{\texttt{wifeOf}} Y$ &
    $\Rightarrow$ &  & .46 & .14 & .06 \\
    \midrule
    $X \xRightarrow{\texttt{brotherOf}} Z \xRightarrow{\texttt{brotherOf}} Y$ &
    $\Rightarrow$ & $X \xRightarrow{\texttt{brotherOf}} Y$ & .86 & .14 & .12 \\
    $X \xRightarrow{\texttt{nephewOf}} Z \xRightarrow{\texttt{uncleOf}} Y$ &
    $\Rightarrow$ &  & .77 & .13 & .10 \\
    $X \xRightarrow{\texttt{brotherOf}} Z \xRightarrow{\texttt{sisterOf}} Y$ &
    $\Rightarrow$ &  & .81 & .13 & .10 \\
    $X \xRightarrow{\texttt{sonOf}} Z \xRightarrow{\texttt{fatherOf}} Y$ &
    $\Rightarrow$ &  & .100 & .08 & .08 \\
    $X \xRightarrow{\texttt{nephewOf}} Z \xRightarrow{\texttt{auntOf}} Y$ &
    $\Rightarrow$ &  & .68 & .11 & .08 \\
    \midrule
    $X \xRightarrow{\texttt{brotherOf}} Z \xRightarrow{\texttt{uncleOf}} Y$ &
    $\Rightarrow$ & $X \xRightarrow{\texttt{uncleOf}} Y$ & .85 & .23 & .20 \\
    $X \xRightarrow{\texttt{uncleOf}} Z \xRightarrow{\texttt{brotherOf}} Y$ &
    $\Rightarrow$ &  & .82 & .22 & .18 \\
    $X \xRightarrow{\texttt{brotherOf}} Z \xRightarrow{\texttt{auntOf}} Y$ &
    $\Rightarrow$ &  & .78 & .22 & .17 \\
    $X \xRightarrow{\texttt{uncleOf}} Z \xRightarrow{\texttt{sisterOf}} Y$ &
    $\Rightarrow$ &  & .74 & .18 & .13 \\
    $X \xRightarrow{\texttt{brotherOf}} Z \xRightarrow{\texttt{fatherOf}} Y$ &
    $\Rightarrow$ &  & .62 & .09 & .06 \\
    $X \xRightarrow{\texttt{brotherOf}} Z \xRightarrow{\texttt{motherOf}} Y$ &
    $\Rightarrow$ &  & .38 & .05 & .02 \\
    \midrule
    $X \xRightarrow{\texttt{nephewOf}} Z \xRightarrow{\texttt{brotherOf}} Y$ &
    $\Rightarrow$ & $X \xRightarrow{\texttt{nephewOf}} Y$ & .86 & .25 & .21 \\
    $X \xRightarrow{\texttt{nephewOf}} Z \xRightarrow{\texttt{sisterOf}} Y$ &
    $\Rightarrow$ &  & .79 & .22 & .17 \\
    $X \xRightarrow{\texttt{brotherOf}} Z \xRightarrow{\texttt{nephewOf}} Y$ &
    $\Rightarrow$ &  & .79 & .21 & .16 \\
    $X \xRightarrow{\texttt{brotherOf}} Z \xRightarrow{\texttt{nieceOf}} Y$ &
    $\Rightarrow$ &  & .72 & .17 & .12 \\
    $X \xRightarrow{\texttt{sonOf}} Z \xRightarrow{\texttt{brotherOf}} Y$ &
    $\Rightarrow$ &  & .64 & .10 & .06 \\
    $X \xRightarrow{\texttt{sonOf}} Z \xRightarrow{\texttt{sisterOf}} Y$ &
    $\Rightarrow$ &  & .36 & .05 & .02 \\
    \bottomrule
\end{longtable}
\end{center}

We use rule \texttt{motherOf}($X$, $Z$) $\wedge$ \texttt{sonOf}($Z$, $Y$) $\Rightarrow$
\texttt{wifeOf}($X$, $Y$) as an example in Table \ref{tab:saturation-family}, where the left
part of the rule is denoted as $\textbf{p}_l$ and the right $q$. This rule can be translated that
if $X$ is mother of $Z$ and $Z$ is son of $Y$, then we can infer that $X$ is wife of $Y$.
The \emph{macro saturation} $\gamma^{\textbf{p}_l}_q = 0.47$ means that
47\% of the factual triplets whose predicate is \texttt{wifeOf}
cover the reasoning rule \texttt{motherOf} $\wedge$ \texttt{sonOf}.
In Table \ref{tab:saturation-family}, $\gamma^{\textbf{p}_l}_q$ roughly tells
the percentage of a potential rule pattern in subgraph $\mathcal{G}(q)$,
whereas the \emph{micro saturation} contains more detailed information focused on one triplet.
$\delta^{\textbf{p}_l}_q = 0.35$ represents that on average, among all rule paths no longer than $L$
that could reason out the predicate \texttt{daughterOf}, more than one third of them follow the pattern
\texttt{motherOf} $\wedge$ \texttt{sonOf}, which is fairly a high proportion.
Finally we heuristically propose \emph{comprehensive saturation} as a global metric
that combines these two factors and may individually serve as a score of a rule
where higher the score indicates more obvious statistical features during inference.

Apart from this, we want to share some more heuristic opinions upon \emph{saturations}.
Firstly we can say the rule \texttt{wifeOf} $\wedge$ \texttt{fatherOf} is \emph{macro-saturated}
with regard to the predicate \texttt{motherOf}, because of its $\gamma^{\textbf{p}_l}_q = 1$.
When saturation of a rule increases, it demonstrates that the rule is more saturated
in comparison to other rules. Secondly, the rules with high saturation
shown in Table \ref{tab:saturation-family} gain distinguished comprehensibility
by human as a reasoning pattern, thus \emph{saturation} may be a valuable complementary
indicator to evaluate the performance and interpretability of a knowledge graph reasoning model.
To the end, during the computation of saturations, we are seemingly in a process of performing
a type of \emph{frequent pattern mining} \cite{han2004mining, agrawal1994fast, dur2000three},
which may be a future ground of research in the area of knowledge graph reasoning.

\textbf{Bifurcation}. Then we report on observations about \emph{bifurcation} which is defined on
a particular predicate. We choose a small group of predicates from several datasets
as illustrated in Table \ref{tab:bifurcations}.
Meanwhile, models in this work are evaluated over queries in form of $(h, q, ?)$,
so that we solely calculate and show the forward bifurcation (\ie $fw\mbox{-}bifur^q(\lambda)$).
We will put more statistics about \emph{bifurcation} in the appendix as well.

Table \ref{tab:bifurcations} shows the intuitive diversity of bifurcation between predicates
within each dataset in multi-target case. First we can pay attention to the predicate \texttt{uncleOf}
in the Family dataset, whose bifurcation with $\lambda=2$ is 84\%, meaning that most of the uncles have
more than two nieces (nephews). The difference between two consecutive numbers in the same row
is also of great value, e.g., the bifurcation with $\lambda=2$ and $\lambda=3$
for the predicate \texttt{daughterOf} is 84\% and 0 respectively,
which means that 84\% of the daughters in the Family dataset have
two parents ($84-0=84$), and none of them have more than two parents ($fw\mbox{-}bifur^q(3)=0$).

\begin{table*}[t]
  \caption{
    Bifurcation(\%) of Family, UMLS and WN-18. The first column lists selected
    datasets and the predicates are shown on the second column.
  }
  \label{tab:bifurcations}
  \centering
  \begin{tabular}{cccccccc}
    \toprule
    &  & \multicolumn{6}{c}{$fw\mbox{-}bifur^q(\lambda)$} \\
    \cmidrule(lr){3-8}
    & {} & $\lambda=2$ & $\lambda=3$ & $\lambda=4$ & $\lambda=5$ & $\lambda=6$ & $\lambda=7$ \\
    \midrule
    \multirow{6}{*}{Family}
    & husbandOf & 14 & 2 & 1 & 0 & 0 & 0 \\
    & wifeOf & 8 & 1 & 0 & 0 & 0 & 0 \\
    & sonOf & 85 & 0 & 0 & 0 & 0 & 0 \\
    & daughterOf & 84 & 0 & 0 & 0 & 0 & 0 \\
    & brotherOf & 77 & 57 & 42 & 30 & 23 & 18 \\
    & uncleOf & 84 & 74 & 64 & 52 & 44 & 39 \\
    \midrule
    \multirow{4}{*}{UMLS}
    & issueIn & 99 & 0 & 0 & 0 & 0 & 0 \\
    & precede & 93 & 93 & 93 & 93 & 50 & 0 \\
    & prevents & 100 & 100 & 100 & 100 & 100 & 40 \\
    & associatedWith & 78 & 64 & 61 & 58 & 56 & 53 \\
    \midrule
    \multirow{4}{*}{WN-18}
    & hasPart & 38 & 21 & 15 & 11 & 9 & 7 \\
    & memberOfDomainUsage & 84 & 52 & 48 & 48 & 44 & 44 \\
    & memberOfDomainTopic & 68 & 58 & 48 & 40 & 35 & 31 \\
    & memberOfDomainRegion & 53 & 35 & 29 & 24 & 21 & 19 \\
    \bottomrule
  \end{tabular}
\end{table*}

\subsubsection{\marktext{Results on knowledge graph completion}}
We evaluate our model in comparison with some baselines\footnotemark on KG completion benchmarks
as stated in Section \ref{subsubsec:datasets} and Section \ref{subsubsec:algorithms}.
Since Neural LP \cite{yang2017differentiable}, DRUM \cite{sadeghian2019drum} and ours all follow
a similar framework, we ensure the same hyperparameter setting during evaluation on these models,
where the maximum rule length $L$ is 2 and the rank of the estimator is $R=3$.
Part of the results are summarized in Table \ref{tab:results-whole}, and more are available in the appendix.

It is clear that our MPLR achieves state-of-the-art results at all metrics
on datasets listed in Table \ref{tab:results-whole} among all methods,
as one can see an obvious improvement on almost all datasets.
Apart from this, our model outperforms Neural LP and DRUM on two real-world datasets shown in the appendix.
We conjecture that this is due to the optimization that enables our model to utilize more training
data at a time and the advancement in multi-target cases.

Notably, it is not fair to compare MPLR with embedding-based methods solely on the
aforementioned metrics, because they are black boxes inside that do not provide interpretability,
while our model has advantages in this area. We will show some of the rules mined by our model later.

\footnotetext{The URLs we use to implement these models are listed in App. \ref{appsec:urls}.}

\begin{table*}[pos=t, width=.75\linewidth]
  \caption{Knowledge graph completion performance comparison. Hit@$k$ is in \%.}
  \label{tab:results-whole}
  \centering
  \begin{tabular*}{\tblwidth}{cccccccccc}
    \toprule
    & \multicolumn{3}{c}{Family} & \multicolumn{3}{c}{Kinship} & \multicolumn{3}{c}{UMLS} \\
    \cmidrule(lr){2-4} \cmidrule(lr){5-7} \cmidrule(lr){8-10}
    & MRR & Hit@1 & Hit@3 & MRR & Hit@1 & Hit@3 & MRR & Hit@1 & Hit@3 \\
    \midrule
    TransE & .14 & 4 & 16 & .10 & 2 & 8 & .13 & 1 & 10 \\
    DistMult & .30 & 11 & 35 & .20 & 5 & 18 & .09 & .6 & 3 \\
    ComplEx & .35 & 15 & 42 & .22 & 7 & 23 & .13 & 9 & 2 \\
    TuckER & .33 & 13 & 39 & .18 & 3 & 15 & .11 & 2 & 6 \\
    RotatE & .41 & 22 & 48 & .26 & 9 & 26 & .15 & 4 & 12 \\
    ConvE & .20 & 8 & 22 & .17 & 2 & 11 & .12 & 1 & 9 \\
    \midrule
    RNNLogic & .27 & 15 & 32 & .28 & 11 & 31 & .21 & 8 & 21 \\
    Neural LP & .50 & 34 & 57 & .25 & 9 & 26 & .26 & 11 & 27 \\
    DRUM & .52 & 35 & 60 & .29 & 12 & 30 & .28 & 14 & 30 \\
    MPLR & \textbf{.64} & \textbf{54} & \textbf{68} & \textbf{.31} & \textbf{16} & \textbf{33} &
      \textbf{.36} & \textbf{25} & \textbf{36} \\
    \bottomrule
  \end{tabular*}
\end{table*}

\begin{table*}[pos=t, width=.79\linewidth]
  \caption{
    Results of reasoning on the Family dataset for specific predicates. Hit@$k$ is in \%.
  }
  \label{tab:results-predicates}
  \centering
  \begin{tabular*}{\tblwidth}{cccccccccc}
    \toprule
    & \multicolumn{3}{c}{husbandOf} & \multicolumn{3}{c}{wifeOf} & \multicolumn{3}{c}{sonOf} \\
    \cmidrule(lr){2-4} \cmidrule(lr){5-7} \cmidrule(lr){8-10}
    & MRR & Hit@1 & Hit@3 & MRR & Hit@1 & Hit@3 & MRR & Hit@1 & Hit@3 \\
    \midrule
    Neural LP \cite{yang2017differentiable} & .49 & 21 & 77 & .48 & 25 & 69 & .76 & 69 & 80 \\
    DRUM \cite{sadeghian2019drum} & .46 & 15 & 77 & .54 & 27 & 69 & .79 & \textbf{69} & 88 \\
    MPLR & \textbf{.78} & \textbf{75} & \textbf{80} & \textbf{.72} & \textbf{70} & \textbf{73}
         & \textbf{.79} & 68 & \textbf{89} \\
    \midrule
    & \multicolumn{3}{c}{daughterOf} & \multicolumn{3}{c}{brotherOf} & \multicolumn{3}{c}{uncleOf} \\
    \cmidrule(lr){2-4} \cmidrule(lr){5-7} \cmidrule(lr){8-10}
    & MRR & Hit@1 & Hit@3 & MRR & Hit@1 & Hit@3 & MRR & Hit@1 & Hit@3 \\
    \midrule
    Neural LP \cite{yang2017differentiable} & .70 & 63 & 71 & .51 & 30 & 62 & .27 & 10 & 28 \\
    DRUM \cite{sadeghian2019drum} & .75 & 65 & \textbf{82} & .54 & 35 & 64 & .42 & 26 & \textbf{47} \\
    MPLR & \textbf{.75} & \textbf{65} & 80 & \textbf{.67} & \textbf{55} & \textbf{71}
         & \textbf{.45} & \textbf{32} & 46 \\
    \bottomrule
  \end{tabular*}
\end{table*}

To demonstrate more details about the capability of models to induce logical rules, we compare our
model against other two models in neural logic programming upon specific predicates.
We choose the Family dataset for better visual availability and
the results are shown in Table \ref{tab:results-predicates}.

Compared with Neural LP and DRUM, our MPLR witnesses a significant improvement on almost all metrics
of predicates. Moreover, during the experiment, we discover that evaluating a reasoning model
simply on Hit@$k$ lacks overallness and precision. The metric Hit@$k$ depends not only on
the performance of model, but also the indicator \emph{bifurcation}. As analysed in Table \ref{tab:bifurcations},
the bifurcations of predicate \texttt{daughterOf} shows that 16\% of the daughters have only one
parent and 84\% of them have two. Thus, Hit@1 of any model should be at most 57\%
($14\% + 84\% / 2$) on the whole KG. In fact, assume all of the daughters have only two parents, \textit{i.e.},
$fw\mbox{-}bifur^q (2)=100\%$ and $fw\mbox{-}bifur^q (3)=0$, then there should be at most one parent
of each daughter ranking the first, and the other not the first, therefore the maximum of Hit@1 is 50\%.

To explain the results shown in Table \ref{tab:results-predicates}, we further compute the bifurcation
on test data of Family, as shown in Table \ref{tab:bifurcations-family-test}.
The maximum Hit@1 of \texttt{daughterOf} on test data should be 97.5\%.
Meanwhile, higher $fw\mbox{-}bifur^q(\lambda)$ empirically means that it is harder to get
a higher Hit@$k$ for $k < \lambda$. The same procedure may be easily adapted to obtain the upper
bound of Hit@$k$ of a model at any knowledge graph completion task.

\begin{table*}[pos=t, width=.33\linewidth]
  \caption{Bifurcation(\%) on test data of Family.}
  \label{tab:bifurcations-family-test}
  \centering
  \begin{tabular*}{\tblwidth}{cccc}
    \toprule
    & \multicolumn{3}{c}{$fw\mbox{-}bifur^q(\lambda)$} \\
    \cmidrule(lr){2-4}
    & $\lambda=2$ & $\lambda=3$ & $\lambda=4$ \\
    \midrule
    husbandOf & 2 & 0 & 0 \\
    wifeOf & 1 & 0 & 0 \\
    sonOf & 3 & 0 & 0 \\
    daughterOf & 5 & 0 & 0 \\
    brotherOf & 23 & 4 & 1 \\
    uncleOf & 40 & 11 & 2 \\
    \bottomrule
  \end{tabular*}
\end{table*}

\begin{table*}[pos=t, width=.45\linewidth]
  \caption{Top rules learned by MPLR on the Family dataset.}
  \label{tab:mined-rules}
  \centering
  \begin{tabular*}{\tblwidth}{rcl}
    \toprule
    Rule & $\Rightarrow$ & Predicate \\
    \midrule
    $X \xRightarrow{\texttt{motherOf}} Z \xRightarrow{\texttt{daughterOf}} Y$ &
    $\Rightarrow$ & $X \xRightarrow{\texttt{wifeOf}} Y$ \\
    $X \xRightarrow{\texttt{motherOf}} Z \xRightarrow{\texttt{sonOf}} Y$ &
    $\Rightarrow$ & \\
    \midrule
    $X \xRightarrow{\texttt{wifeOf}} Z \xRightarrow{\texttt{fatherOf}} Y$ &
    $\Rightarrow$ & $X \xRightarrow{\texttt{motherOf}} Y$ \\
    $X \xRightarrow{\texttt{motherOf}} Z \xRightarrow{\texttt{sisterOf}} Y$ &
    $\Rightarrow$ & \\
    \textcolor{red}{$X \xRightarrow{\texttt{wifeOf}} Z \xRightarrow{\texttt{wifeOf}} Y$} &
    $\Rightarrow$ & \\
    $X \xRightarrow{\texttt{motherOf}} Z \xRightarrow{\texttt{brotherOf}} Y$ &
    $\Rightarrow$ & \\
    \midrule
    $X \xRightarrow{\texttt{brotherOf}} Z \xRightarrow{\texttt{fatherOf}} Y$ &
    $\Rightarrow$ & $X \xRightarrow{\texttt{uncleOf}} Y$ \\
    $X \xRightarrow{\texttt{uncleOf}} Z \xRightarrow{\texttt{brotherOf}} Y$ &
    $\Rightarrow$ & \\
    $X \xRightarrow{\texttt{brotherOf}} Z \xRightarrow{\texttt{motherOf}} Y$ &
    $\Rightarrow$ & \\
    $X \xRightarrow{\texttt{motherOf}} Z \xRightarrow{\texttt{brotherOf}} Y$ &
    $\Rightarrow$ & \\
    \textcolor{red}{$X \xRightarrow{\texttt{sisterOf}} Z \xRightarrow{\texttt{brotherOf}} Y$} &
    $\Rightarrow$ & \\
    $X \xRightarrow{\texttt{uncleOf}} Z \xRightarrow{\texttt{sisterOf}} Y$ &
    $\Rightarrow$ & \\
    \bottomrule
  \end{tabular*}
\end{table*}

\subsection{Experiment on interpretability of mined rules}

Neural LP framework successfully combines structure learning and parameter learning. It not only
induces multiple logical rules to capture the complex structure in the knowledge graph, but also
learns to distribute confidences on rules \cite{yang2017differentiable}. In addition to
the evolution on KG completion task in multi-target situation, our model also succeeds
Neural LP on interpretability and further, becomes more interpretable to human.
Throughout this section we use the Family dataset for visualization purposes as it is more tangible.
Other datasets like UMLS produce similar outputs.

We sort the rules generated by MPLR according to their assigned confidences
and show top rules in Table \ref{tab:mined-rules}. To be honest, because of the constraints
on expressiveness, there are logically incorrect rules mined by this model, which is highlighted
by \textcolor{red}{red color} in the table. We will explain this in the next section. For more
learned logical rules, please refer to App. \ref{appsec:rules}.

We can see the rules are of high quality and of good diversity, although there are few inappropriate ones.
More importantly, the mined rules shown in Table \ref{tab:mined-rules} reaches a great agreement
with high-saturated rules in Table \ref{tab:saturation-family}, which indeed reflects the power
of \emph{reasoning saturation} as an indicator and on the other hand depicts the strong interpretability of MPLR.

\subsection{Discussion on Neural LP framework}

\begin{figure}[t]
  \centering
  \includegraphics[width=0.5\linewidth]{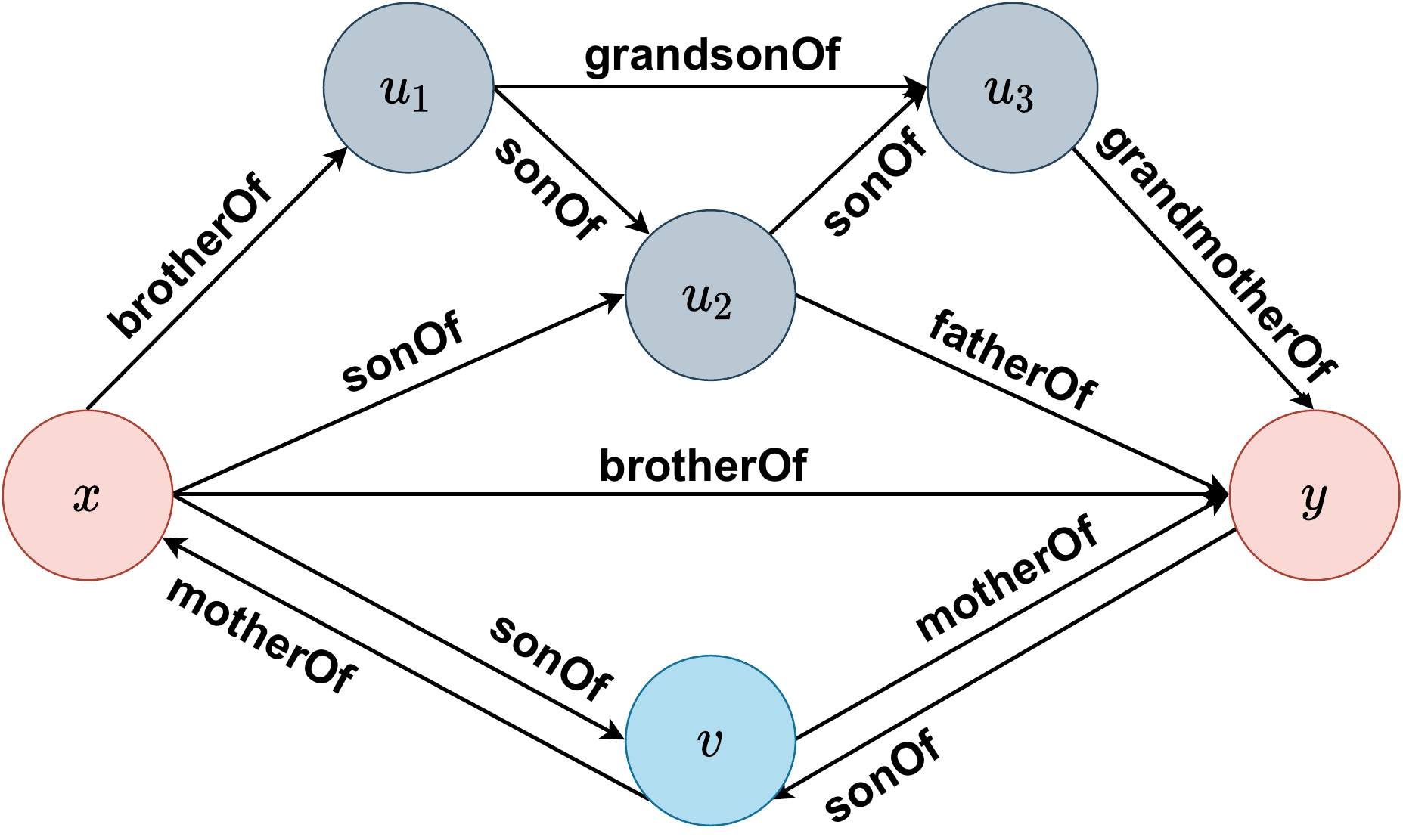}
  \caption{
    An example to demonstrate the dilemma of existing KG reasoning models
    to effectively induce among rules of different lengths. \texttt{brotherOf}($x$, $y$)
    is the query in this example, and there are multiple rules that guide a way from $x$
    to the answer $y$.
  }
  \label{fig:multi-hop}
\end{figure}

Despite the fact that Neural LP is an end-to-end gradient-based KG reasoning framework and
fills in the gap between traditional KG reasoning models (\eg embedding methods)
and interpretability, through close observation on datasets and analysis on formulas,
we discover that there also exist some restrictions for Neural LP-based algorithms:

\begin{enumerate}
  \item As proved in \cite{sadeghian2019drum}, current framework inevitably mines incorrect rules
    with high confidences, \textit{i.e.}, if there are several rules sharing one or more predicates,
    confidences of rules would be coupled mutually. Intuitively, this is because Eq. (\ref{eq:phi})
    distributes the score of a rule to the predicates that constitute the rule at different hops.
    For instance, \texttt{brotherOf} $\wedge$ \texttt{sonOf} and \texttt{brotherOf}
    $\wedge$ \texttt{sisterOf} share \texttt{brotherOf} at first hop.
    However, in case our query is \texttt{sonOf}, if \texttt{brotherOf} wins high confidence at first hop,
    the score of the second rule may not be too low, which is absolutely an incorrect result.
    This reduces the interpretability of the output rules.
  \item Present models are faced with the dilemma where there would be invertible relation
     pairs and rules of varied lengths mixing up. \emph{(i)} A relation pair $(R_1, R_2)$
     is invertible if there simultaneously exist two triplets $(h, R_1, t)$ and $(t, R_2, h)$
     in a KG. \emph{(ii)} The KG example shown in Fig. \ref{fig:multi-hop} consists of
     candidate rules of length 2, 3 and 4 for query \texttt{brotherOf}($x$, $y$).
     These two factors jointly may cause invalid induction results, under the condition that
     we choose an improper hyperparameter $L$ as the maximum length of rules, e.g., if we set $L=4$,
     the rule path \texttt{sonOf}($x$, $v$) $\wedge$ \texttt{motherOf}($v$, $x$) $\wedge$
     \texttt{sonOf}($x$, $v$) $\wedge$ \texttt{motherOf}($v$, $y$) $\Rightarrow$
     \texttt{brotherOf}($x$, $y$) is possible but  meaningless. This is also
     essentially due to the distribution of confidences brought by Eq. (\ref{eq:phi}),
     and thus impedes the way for multi-hop reasoning over long rules.
  \item A high ranking of an entity results not solely from a top-scored rule,
    but also a number of relatively low-scored rules. As formulated in Section \ref{subsec:neurallp},
    the product of vector-matrix multiplication is a scalar representing the number of unique paths,
    and the final score of the entity is computed by summing up the confidences of all paths.
    Again, metrics like MRR and Hit@$k$ only assess models in terms of the ranking of the desired entity,
    rather than the quality of mined reasoning rules.
    Thus, models following the Neural LP framework with high MRR and Hit@$k$ may be better suited to
    tasks like question answering or relation completion \cite{ji2021survey},
    but this may not be applied to rule mining.
\end{enumerate}

\section{Conclusions}\label{sec:conclusion}

In this paper, we firstly propose novel indicators that help to understand knowledge graph
reasoning tasks and to serve as a supplement to the existing metrics (\eg Hit@$k$) for evaluating models.
The \emph{saturation} measures the possibility of a rule being a plausible inference for a relation,
which fills in the blanks of judging the interpretability of mined rules.
While the \emph{bifurcation}, computing the proportion of instances with multiple reasoning destinations,
is useful for enhancing the power of MMR and Hit@$k$. Then we address the problem of learning rules
from knowledge graphs in multi-target cases where a model called MPLR is proposed.
MPLR improves the Neural LP framework in order to allow more queries fed in one batch,
thus fits in multi-target scenarios. Experiment results have shown that our proposed method
improves performance on several knowledge graph reasoning datasets and owns strong interpretability,
under the evaluation of traditional metrics and our newly suggested ones. In the future,
we would like to break the limitation of our current model for multi-hop reasoning where the rules
are much longer.

\section*{Competing Interests}
We declare no competing interests.

\section*{Acknowledgment}
The work of this paper is supported by the "National Key R\&D Program of China" (2020YFB2009502),
"the Fundamental Research Funds for the Central Universities" (Grant No. HIT.NSRIF.2020098).
\clearpage

\appendix
\section{Extension to table \ref{tab:saturation-family}: saturations of UMLS}

\begin{table*}[pos=h, width=0.7\linewidth]
  \caption{
    Saturations of UMLS (without sampling).
    The rule length is fixed to 2. $\gamma^{\textbf{p}_l}_q$, $\delta^{\textbf{p}_l}_q$,
    $\eta^{\textbf{p}_l}_q$ are \emph{macro}, \emph{micro} and \emph{comprehensive} saturations.
    The results are sorted by the comprehensive saturation in descending order.
  }
  \centering
  \begin{tabular*}{\tblwidth}{rclccc}
    \toprule
    Rule & $\Rightarrow$ & Predicate & $\gamma^{\textbf{p}_l}_q$ &
      $\delta^{\textbf{p}_l}_q$ & $\eta^{\textbf{p}_l}_q$ \\
    \midrule
    $X \xRightarrow{\texttt{manifestationOf}} Z \xRightarrow{\texttt{resultOf}} Y$ &
    $\Rightarrow$ & $X \xRightarrow{\texttt{manifestationOf}} Y$ & 1. & .05 & .05 \\
    $X \xRightarrow{\texttt{manifestationOf}} Z \xRightarrow{\texttt{affects}} Y$ &
    $\Rightarrow$ &  & .91 & .04 & .04 \\
    $X \xRightarrow{\texttt{manifestationOf}} Z \xRightarrow{\texttt{processOf}} Y$ &
    $\Rightarrow$ &  & .91 & .04 & .04 \\
    $X \xRightarrow{\texttt{resultOf}} Z \xRightarrow{\texttt{resultOf}} Y$ &
    $\Rightarrow$ &  & .71 & .05 & .04 \\
    $X \xRightarrow{\texttt{resultOf}} Z \xRightarrow{\texttt{affects}} Y$ &
    $\Rightarrow$ &  & .75 & .04 & .03 \\
    \midrule
    $X \xRightarrow{\texttt{interactWith}} Z \xRightarrow{\texttt{performs}} Y$ &
    $\Rightarrow$ & $X \xRightarrow{\texttt{performs}} Y$ & .83 & .31 & .26 \\
    $X \xRightarrow{\texttt{isA}} Z \xRightarrow{\texttt{performs}} Y$ &
    $\Rightarrow$ &  & .83 & .13 & .11 \\
    $X \xRightarrow{\texttt{performs}} Z \xRightarrow{\texttt{isA}} Y$ &
    $\Rightarrow$ &  & .33 & .16 & .05 \\
    \midrule
    $X \xRightarrow{\texttt{interactWith}} Z \xRightarrow{\texttt{ingredientOf}} Y$ &
    $\Rightarrow$ & $X \xRightarrow{\texttt{ingredientOf}} Y$ & .96 & .61 & .52 \\
    $X \xRightarrow{\texttt{isA}} Z \xRightarrow{\texttt{ingredientOf}} Y$ &
    $\Rightarrow$ &  & .86 & .32 & .31 \\
    \midrule
    $X \xRightarrow{\texttt{interactWith}} Z \xRightarrow{\texttt{exhibits}} Y$ &
    $\Rightarrow$ & $X \xRightarrow{\texttt{exhibits}} Y$ & .87 & .29 & .25 \\
    $X \xRightarrow{\texttt{exhibits}} Z \xRightarrow{\texttt{affects}} Y$ &
    $\Rightarrow$ &  & 1. & .21 & .21 \\
    $X \xRightarrow{\texttt{isA}} Z \xRightarrow{\texttt{exhibits}} Y$ &
    $\Rightarrow$ &  & .87 & .15 & .13 \\
    $X \xRightarrow{\texttt{performs}} Z \xRightarrow{\texttt{affects}} Y$ &
    $\Rightarrow$ &  & .40 & .07 & .03 \\
    \bottomrule
  \end{tabular*}
\end{table*}

\section{\marktext{Model URLs}}\label{appsec:urls}

The models we use are available at the following URLs:
\begin{itemize}
  \item TransE, DistMult and ComplEx \cite{ampligraph} (2019)
  \item TuckER \cite{tuckergithub} (2019)
  \item RotatE \cite{rotategithub} (2018)
  \item ConvE \cite{convegithub} (2018)
  \item RNNLogic \cite{rnnlogicgithub} (2021)
  \item Neural LP \cite{neurallpgithub} (2017)
  \item DRUM \cite{drumgithub} (2019)
  \item MPLR (ours) \cite{mplrgithub} (2021)
\end{itemize}
\clearpage

\section{Extension to table \ref{tab:bifurcations}: bifurcation of all datasets}

\begin{table*}[pos=h, width=.7\linewidth]
  \caption{
    Bifurcation(\%) of all datasets. The first column lists selected
    datasets and the predicates are shown on the second column.
  }
  \centering
  \begin{tabular*}{\tblwidth}{cccccccc}
    \toprule
    &  & \multicolumn{6}{c}{$fw\mbox{-}bifur^q(\lambda)$} \\
    \cmidrule(lr){3-8}
    & {} & $\lambda=2$ & $\lambda=3$ & $\lambda=4$ & $\lambda=5$ & $\lambda=6$ & $\lambda=7$ \\
    \midrule
    \multirow{8}{*}{FB15K-237}
    & position & 94 & 92 & 92 & 92 & 91 & 89 \\
    & nominatedFor & 93 & 89 & 86 & 81 & 79 & 76 \\
    & awardWinner & 74 & 59 & 48 & 41 & 33 & 29 \\
    & award & 66 & 46 & 33 & 25 & 19 & 14 \\
    & list & 17 & 11 & 0 & 0 & 0 & 0 \\
    & participant & 16 & 6 & 2 & 1 & 1 & 0 \\
    & season & 1 & 1 & 1 & 94 & 90 & 90 \\
    & artist & 67 & 67 & 67 & 67 & 67 & 67 \\
    \midrule
    \multirow{4}{*}{WN18}
    & alsoSee & 48 & 23 & 11 & 6 & 2 & 1 \\
    & hypernym & 2 & 0 & 0 & 0 & 0 & 0 \\
    & hyponym & 56 & 35 & 24 & 18 & 14 & 11 \\
    & partOf & 16 & 4 & 1 & 0 & 0 & 0 \\
    \midrule
    \multirow{6}{*}{Family}
    & auntOf & 85 & 74 & 65 & 54 & 49 & 44 \\
    & fatherOf & 42 & 26 & 17 & 12 & 07 & 05 \\
    & motherOf & 53 & 35 & 22 & 14 & 08 & 06 \\
    & nephewOf & 82 & 70 & 55 & 43 & 34 & 28 \\
    & nieceOf & 87 & 73 & 60 & 49 & 41 & 36 \\
    & sisterOf & 82 & 65 & 52 & 35 & 29 & 24 \\
    \midrule
    \multirow{6}{*}{UMLS}
    & manifestationOf & 100 & 82 & 82 & 82 & 82 & 82 \\
    & evaluationOf & 100 & 100 & 100 & 100 & 100 & 100 \\
    & performs & 100 & 100 & 100 & 100 & 100 & 100 \\
    & ingredientOf & 0 & 0 & 0 & 0 & 0 & 0 \\
    & interactWith & 93 & 87 & 80 & 73 & 67 & 62 \\
    & resultOf & 60 & 57 & 57 & 57 & 57 & 57 \\
    \midrule
    \multirow{5}{*}{Kinship}
    & term25 & 67 & 33 & 0 & 0 & 0 & 0 \\
    & term22 & 69 & 55 & 43 & 33 & 31 & 27 \\
    & term19 & 50 & 50 & 50 & 50 & 25 & 0 \\
    & term18 & 91 & 76 & 61 & 49 & 42 & 38 \\
    & term14 & 75 & 67 & 58 & 42 & 8 & 8 \\
    \bottomrule
  \end{tabular*}
\end{table*}
\clearpage

\section{Extension to table \ref{tab:results-whole}: results on FB15K-237 and WN18}

\begin{table*}[pos=h, width=.55\linewidth]
  \caption{
    Knowledge graph completion results on FB15K-237 and WN18. Hit@$k$ is in \%.
    }
  \centering
  \begin{tabular*}{\tblwidth}{ccccccc}
    \toprule
    & \multicolumn{3}{c}{FB15K-237} & \multicolumn{3}{c}{WN18} \\
    \cmidrule(lr){2-4} \cmidrule(lr){5-7}
    & MRR & Hit@1 & Hit@3 & MRR & Hit@1 & Hit@3 \\
    \midrule
    TransE & .21 & 12 & 23 & .23 & 2 & 37 \\
    DistMult & .24 & 15 & 26 & .53 & 39 & 62 \\
    ComplEx & .23 & 14 & 25 & \textbf{.60} & \textbf{47} & \textbf{67} \\
    TuckER & .25 & \textbf{18} & \textbf{26} & .57 & 45 & 65 \\
    RotatE & .11 & 5 & 11 & .22 & 15 & 24 \\
    ConvE & \textbf{.27} & 16 & 23 & .22 & 15 & 22 \\
    \midrule
    RNNLogic & .11 & 8 & 11 & .01 & 1 & 1 \\
    Neural LP & .20 & 15 & 22 & .13 & 2 & 13 \\
    DRUM & .20 & 15 & 22 & .13 & 4 & 13 \\
    MPLR & .20 & 16 & 23 & .16 & 9 & 15 \\
    \bottomrule
  \end{tabular*}
\end{table*}

\section{Extension to table \ref{tab:mined-rules}: more mined rules from the Family dataset}
\label{appsec:rules}
\begin{table*}[pos=h, width=.45\linewidth]
  \caption{
    Rules learned by MPLR on the Family dataset. Rules are sorted by their confidences
    in descending order.
  }
  \centering
  \begin{tabular*}{\tblwidth}{rcl}
    \toprule
    Rule & $\Rightarrow$ & Predicate \\
    \midrule
    $X \xRightarrow{\texttt{sisterOf}} Z \xRightarrow{\texttt{sonOf}} Y$ &
    $\Rightarrow$ & $X \xRightarrow{\texttt{daughterOf}} Y$ \\
    $X \xRightarrow{\texttt{sisterOf}} Z \xRightarrow{\texttt{daughterOf}} Y$ &
    $\Rightarrow$ & \\
    $X \xRightarrow{\texttt{daughterOf}} Z \xRightarrow{\texttt{wifeOf}} Y$ &
    $\Rightarrow$ & \\
    \midrule
    $X \xRightarrow{\texttt{brotherOf}} Z \xRightarrow{\texttt{nephewOf}} Y$ &
    $\Rightarrow$ & $X \xRightarrow{\texttt{nephewOf}} Y$ \\
    $X \xRightarrow{\texttt{brotherOf}} Z \xRightarrow{\texttt{nieceOf}} Y$ &
    $\Rightarrow$ & \\
    \midrule
    $X \xRightarrow{\texttt{sisterOf}} Z \xRightarrow{\texttt{sisterOf}} Y$ &
    $\Rightarrow$ & $X \xRightarrow{\texttt{sisterOf}} Y$ \\
    $X \xRightarrow{\texttt{daughterOf}} Z \xRightarrow{\texttt{motherOf}} Y$ &
    $\Rightarrow$ & \\
    $X \xRightarrow{\texttt{sisterOf}} Z \xRightarrow{\texttt{brotherOf}} Y$ &
    $\Rightarrow$ & \\
    \midrule
    $X \xRightarrow{\texttt{sisterOf}} Z \xRightarrow{\texttt{motherOf}} Y$ &
    $\Rightarrow$ & $X \xRightarrow{\texttt{auntOf}} Y$ \\
    $X \xRightarrow{\texttt{auntOf}} Z \xRightarrow{\texttt{sisterOf}} Y$ &
    $\Rightarrow$ & \\
    \textcolor{red}{$X \xRightarrow{\texttt{fatherOf}} Z \xRightarrow{\texttt{fatherOf}} Y$} &
    $\Rightarrow$ & \\
    $X \xRightarrow{\texttt{sisterOf}} Z \xRightarrow{\texttt{fatherOf}} Y$ &
    $\Rightarrow$ & \\
    \bottomrule
  \end{tabular*}
\end{table*}
\clearpage

\bibliographystyle{cas-model2-names}
\bibliography{bibliography}

\begin{thebibliography}{56}
\expandafter\ifx\csname natexlab\endcsname\relax\def\natexlab#1{#1}\fi
\providecommand{\url}[1]{\texttt{#1}}
\providecommand{\href}[2]{#2}
\providecommand{\path}[1]{#1}
\providecommand{\DOIprefix}{doi:}
\providecommand{\ArXivprefix}{arXiv:}
\providecommand{\URLprefix}{URL: }
\providecommand{\Pubmedprefix}{pmid:}
\providecommand{\doi}[1]{\href{http://dx.doi.org/#1}{\path{#1}}}
\providecommand{\Pubmed}[1]{\href{pmid:#1}{\path{#1}}}
\providecommand{\bibinfo}[2]{#2}
\ifx\xfnm\relax \def\xfnm[#1]{\unskip,\space#1}\fi
\bibitem[{Agrawal et~al.(1994)Agrawal, Srikant et~al.}]{agrawal1994fast}
\bibinfo{author}{Agrawal, R.}, \bibinfo{author}{Srikant, R.}, et~al.,
  \bibinfo{year}{1994}.
\newblock \bibinfo{title}{Fast algorithms for mining association rules}, in:
  \bibinfo{booktitle}{Proc. 20th int. conf. very large data bases, VLDB},
  \bibinfo{organization}{Citeseer}. pp. \bibinfo{pages}{487--499}.
\bibitem[{Bala\v{z}evi\'c et~al.(2019)Bala\v{z}evi\'c, Allen and
  Hospedales}]{tuckergithub}
\bibinfo{author}{Bala\v{z}evi\'c, I.}, \bibinfo{author}{Allen, C.},
  \bibinfo{author}{Hospedales, T.M.}, \bibinfo{year}{2019}.
\newblock \bibinfo{title}{Tucker}.
\newblock \bibinfo{howpublished}{\url{https://github.com/ibalazevic/TuckER}}.
\bibitem[{Bala{\v{z}}evi{\'c} et~al.(2019)Bala{\v{z}}evi{\'c}, Allen and
  Hospedales}]{balavzevic2019tucker}
\bibinfo{author}{Bala{\v{z}}evi{\'c}, I.}, \bibinfo{author}{Allen, C.},
  \bibinfo{author}{Hospedales, T.M.}, \bibinfo{year}{2019}.
\newblock \bibinfo{title}{Tucker: Tensor factorization for knowledge graph
  completion}.
\newblock \bibinfo{journal}{arXiv preprint arXiv:1901.09590} .
\bibitem[{Bollacker et~al.(2008)Bollacker, Evans, Paritosh, Sturge and
  Taylor}]{bollacker2008freebase}
\bibinfo{author}{Bollacker, K.}, \bibinfo{author}{Evans, C.},
  \bibinfo{author}{Paritosh, P.}, \bibinfo{author}{Sturge, T.},
  \bibinfo{author}{Taylor, J.}, \bibinfo{year}{2008}.
\newblock \bibinfo{title}{Freebase: a collaboratively created graph database
  for structuring human knowledge}, in: \bibinfo{booktitle}{Proceedings of the
  2008 ACM SIGMOD international conference on Management of data}, pp.
  \bibinfo{pages}{1247--1250}.
\bibitem[{Bordes et~al.(2013)Bordes, Usunier, Garcia-Duran, Weston and
  Yakhnenko}]{bordes2013translating}
\bibinfo{author}{Bordes, A.}, \bibinfo{author}{Usunier, N.},
  \bibinfo{author}{Garcia-Duran, A.}, \bibinfo{author}{Weston, J.},
  \bibinfo{author}{Yakhnenko, O.}, \bibinfo{year}{2013}.
\newblock \bibinfo{title}{Translating embeddings for modeling multi-relational
  data}.
\newblock \bibinfo{journal}{Advances in neural information processing systems}
  \bibinfo{volume}{26}.
\bibitem[{Cohen(2016)}]{cohen2016tensorlog}
\bibinfo{author}{Cohen, W.W.}, \bibinfo{year}{2016}.
\newblock \bibinfo{title}{Tensorlog: A differentiable deductive database}.
\newblock \bibinfo{journal}{arXiv preprint arXiv:1605.06523} .
\bibitem[{Costabello et~al.(2019)Costabello, Pai, Van, McGrath, McCarthy and
  Tabacof}]{ampligraph}
\bibinfo{author}{Costabello, L.}, \bibinfo{author}{Pai, S.},
  \bibinfo{author}{Van, C.L.}, \bibinfo{author}{McGrath, R.},
  \bibinfo{author}{McCarthy, N.}, \bibinfo{author}{Tabacof, P.},
  \bibinfo{year}{2019}.
\newblock \bibinfo{title}{{AmpliGraph: a Library for Representation Learning on
  Knowledge Graphs}}.
\newblock \URLprefix \url{https://doi.org/10.5281/zenodo.2595043},
  \DOIprefix\doi{10.5281/zenodo.2595043}.
\bibitem[{Dettmers et~al.(2018a)Dettmers, Minervini, Stenetorp and
  Riedel}]{dettmers2018convolutional}
\bibinfo{author}{Dettmers, T.}, \bibinfo{author}{Minervini, P.},
  \bibinfo{author}{Stenetorp, P.}, \bibinfo{author}{Riedel, S.},
  \bibinfo{year}{2018}a.
\newblock \bibinfo{title}{Convolutional 2d knowledge graph embeddings}, in:
  \bibinfo{booktitle}{Thirty-second AAAI conference on artificial
  intelligence}.
\bibitem[{Dettmers et~al.(2018b)Dettmers, Pasquale, Pontus and
  Riedel}]{convegithub}
\bibinfo{author}{Dettmers, T.}, \bibinfo{author}{Pasquale, M.},
  \bibinfo{author}{Pontus, S.}, \bibinfo{author}{Riedel, S.},
  \bibinfo{year}{2018}b.
\newblock \bibinfo{title}{Conve}.
\newblock \bibinfo{howpublished}{\url{https://github.com/TimDettmers/ConvE}}.
\bibitem[{Ding et~al.(2019)Ding, Zhou, Chen, Yang and Tang}]{ding2019cognitive}
\bibinfo{author}{Ding, M.}, \bibinfo{author}{Zhou, C.}, \bibinfo{author}{Chen,
  Q.}, \bibinfo{author}{Yang, H.}, \bibinfo{author}{Tang, J.},
  \bibinfo{year}{2019}.
\newblock \bibinfo{title}{Cognitive graph for multi-hop reading comprehension
  at scale}.
\newblock \bibinfo{journal}{arXiv preprint arXiv:1905.05460} .
\bibitem[{Dong et~al.(2014)Dong, Gabrilovich, Heitz, Horn, Lao, Murphy,
  Strohmann, Sun and Zhang}]{dong2014knowledge}
\bibinfo{author}{Dong, X.}, \bibinfo{author}{Gabrilovich, E.},
  \bibinfo{author}{Heitz, G.}, \bibinfo{author}{Horn, W.},
  \bibinfo{author}{Lao, N.}, \bibinfo{author}{Murphy, K.},
  \bibinfo{author}{Strohmann, T.}, \bibinfo{author}{Sun, S.},
  \bibinfo{author}{Zhang, W.}, \bibinfo{year}{2014}.
\newblock \bibinfo{title}{Knowledge vault: A web-scale approach to
  probabilistic knowledge fusion}, in: \bibinfo{booktitle}{Proceedings of the
  20th ACM SIGKDD international conference on Knowledge discovery and data
  mining}, pp. \bibinfo{pages}{601--610}.
\bibitem[{D{\"u}r et~al.(2000)D{\"u}r, Vidal and Cirac}]{dur2000three}
\bibinfo{author}{D{\"u}r, W.}, \bibinfo{author}{Vidal, G.},
  \bibinfo{author}{Cirac, J.I.}, \bibinfo{year}{2000}.
\newblock \bibinfo{title}{Three qubits can be entangled in two inequivalent
  ways}.
\newblock \bibinfo{journal}{Physical Review A} \bibinfo{volume}{62},
  \bibinfo{pages}{062314}.
\bibitem[{Faudree et~al.(2011)Faudree, Faudree and Schmitt}]{faudree2011survey}
\bibinfo{author}{Faudree, J.R.}, \bibinfo{author}{Faudree, R.J.},
  \bibinfo{author}{Schmitt, J.R.}, \bibinfo{year}{2011}.
\newblock \bibinfo{title}{A survey of minimum saturated graphs}.
\newblock \bibinfo{journal}{The Electronic Journal of Combinatorics}
  \bibinfo{volume}{1000}, \bibinfo{pages}{DS19--Jul}.
\bibitem[{Guu et~al.(2015)Guu, Miller and Liang}]{guu2015traversing}
\bibinfo{author}{Guu, K.}, \bibinfo{author}{Miller, J.},
  \bibinfo{author}{Liang, P.}, \bibinfo{year}{2015}.
\newblock \bibinfo{title}{Traversing knowledge graphs in vector space}.
\newblock \bibinfo{journal}{arXiv preprint arXiv:1506.01094} .
\bibitem[{Hajnal(1965)}]{hajnal1965theorem}
\bibinfo{author}{Hajnal, A.}, \bibinfo{year}{1965}.
\newblock \bibinfo{title}{A theorem on k-saturated graphs}.
\newblock \bibinfo{journal}{Canadian Journal of Mathematics}
  \bibinfo{volume}{17}, \bibinfo{pages}{720--724}.
\bibitem[{Han et~al.(2004)Han, Pei, Yin and Mao}]{han2004mining}
\bibinfo{author}{Han, J.}, \bibinfo{author}{Pei, J.}, \bibinfo{author}{Yin,
  Y.}, \bibinfo{author}{Mao, R.}, \bibinfo{year}{2004}.
\newblock \bibinfo{title}{Mining frequent patterns without candidate
  generation: A frequent-pattern tree approach}.
\newblock \bibinfo{journal}{Data mining and knowledge discovery}
  \bibinfo{volume}{8}, \bibinfo{pages}{53--87}.
\bibitem[{Hochreiter and Schmidhuber(1997)}]{hochreiter1997long}
\bibinfo{author}{Hochreiter, S.}, \bibinfo{author}{Schmidhuber, J.},
  \bibinfo{year}{1997}.
\newblock \bibinfo{title}{Long short-term memory}.
\newblock \bibinfo{journal}{Neural computation} \bibinfo{volume}{9},
  \bibinfo{pages}{1735--1780}.
\bibitem[{Ji et~al.(2021)Ji, Pan, Cambria, Marttinen and Philip}]{ji2021survey}
\bibinfo{author}{Ji, S.}, \bibinfo{author}{Pan, S.}, \bibinfo{author}{Cambria,
  E.}, \bibinfo{author}{Marttinen, P.}, \bibinfo{author}{Philip, S.Y.},
  \bibinfo{year}{2021}.
\newblock \bibinfo{title}{A survey on knowledge graphs: Representation,
  acquisition, and applications}.
\newblock \bibinfo{journal}{IEEE Transactions on Neural Networks and Learning
  Systems} .
\bibitem[{Kathryn and Mazaitis(2018)}]{kathryn2018tensorlog}
\bibinfo{author}{Kathryn, W.W.C.F.Y.}, \bibinfo{author}{Mazaitis, R.},
  \bibinfo{year}{2018}.
\newblock \bibinfo{title}{Tensorlog: Deep learning meets probabilistic
  databases}.
\newblock \bibinfo{journal}{Journal of Artificial Intelligence Research}
  \bibinfo{volume}{1}, \bibinfo{pages}{1--15}.
\bibitem[{Kingma and Ba(2014)}]{kingma2014adam}
\bibinfo{author}{Kingma, D.P.}, \bibinfo{author}{Ba, J.}, \bibinfo{year}{2014}.
\newblock \bibinfo{title}{Adam: A method for stochastic optimization}.
\newblock \bibinfo{journal}{arXiv preprint arXiv:1412.6980} .
\bibitem[{Kok and Domingos(2007)}]{kok2007statistical}
\bibinfo{author}{Kok, S.}, \bibinfo{author}{Domingos, P.},
  \bibinfo{year}{2007}.
\newblock \bibinfo{title}{Statistical predicate invention}, in:
  \bibinfo{booktitle}{Proceedings of the 24th international conference on
  Machine learning}, pp. \bibinfo{pages}{433--440}.
\bibitem[{Koller et~al.(2007)Koller, Friedman, D{\v{z}}eroski, Sutton,
  McCallum, Pfeffer, Abbeel, Wong, Meek, Neville
  et~al.}]{koller2007introduction}
\bibinfo{author}{Koller, D.}, \bibinfo{author}{Friedman, N.},
  \bibinfo{author}{D{\v{z}}eroski, S.}, \bibinfo{author}{Sutton, C.},
  \bibinfo{author}{McCallum, A.}, \bibinfo{author}{Pfeffer, A.},
  \bibinfo{author}{Abbeel, P.}, \bibinfo{author}{Wong, M.F.},
  \bibinfo{author}{Meek, C.}, \bibinfo{author}{Neville, J.}, et~al.,
  \bibinfo{year}{2007}.
\newblock \bibinfo{title}{Introduction to statistical relational learning}.
\newblock \bibinfo{publisher}{MIT press}.
\bibitem[{Lao and Cohen(2010)}]{lao2010relational}
\bibinfo{author}{Lao, N.}, \bibinfo{author}{Cohen, W.W.}, \bibinfo{year}{2010}.
\newblock \bibinfo{title}{Relational retrieval using a combination of
  path-constrained random walks}.
\newblock \bibinfo{journal}{Machine learning} \bibinfo{volume}{81},
  \bibinfo{pages}{53--67}.
\bibitem[{Lin et~al.(2019)Lin, Chen, Chen and Ren}]{lin2019kagnet}
\bibinfo{author}{Lin, B.Y.}, \bibinfo{author}{Chen, X.}, \bibinfo{author}{Chen,
  J.}, \bibinfo{author}{Ren, X.}, \bibinfo{year}{2019}.
\newblock \bibinfo{title}{Kagnet: Knowledge-aware graph networks for
  commonsense reasoning}.
\newblock \bibinfo{journal}{arXiv preprint arXiv:1909.02151} .
\bibitem[{Lin et~al.(2015)Lin, Liu, Sun, Liu and Zhu}]{lin2015learning}
\bibinfo{author}{Lin, Y.}, \bibinfo{author}{Liu, Z.}, \bibinfo{author}{Sun,
  M.}, \bibinfo{author}{Liu, Y.}, \bibinfo{author}{Zhu, X.},
  \bibinfo{year}{2015}.
\newblock \bibinfo{title}{Learning entity and relation embeddings for knowledge
  graph completion}, in: \bibinfo{booktitle}{Twenty-ninth AAAI conference on
  artificial intelligence}.
\bibitem[{Miller(1995)}]{miller1995wordnet}
\bibinfo{author}{Miller, G.A.}, \bibinfo{year}{1995}.
\newblock \bibinfo{title}{Wordnet: a lexical database for english}.
\newblock \bibinfo{journal}{Communications of the ACM} \bibinfo{volume}{38},
  \bibinfo{pages}{39--41}.
\bibitem[{Miller(1998)}]{miller1998wordnet}
\bibinfo{author}{Miller, G.A.}, \bibinfo{year}{1998}.
\newblock \bibinfo{title}{WordNet: An electronic lexical database}.
\newblock \bibinfo{publisher}{MIT press}.
\bibitem[{Muggleton and De~Raedt(1994)}]{muggleton1994inductive}
\bibinfo{author}{Muggleton, S.}, \bibinfo{author}{De~Raedt, L.},
  \bibinfo{year}{1994}.
\newblock \bibinfo{title}{Inductive logic programming: Theory and methods}.
\newblock \bibinfo{journal}{The Journal of Logic Programming}
  \bibinfo{volume}{19}, \bibinfo{pages}{629--679}.
\bibitem[{MYCIN(1976)}]{mycin1976computer}
\bibinfo{author}{MYCIN, E.S.}, \bibinfo{year}{1976}.
\newblock \bibinfo{title}{Computer-based medical consultations}.
\bibitem[{Newell et~al.(1959)Newell, Shaw and Simon}]{newell1959report}
\bibinfo{author}{Newell, A.}, \bibinfo{author}{Shaw, J.C.},
  \bibinfo{author}{Simon, H.A.}, \bibinfo{year}{1959}.
\newblock \bibinfo{title}{Report on a general problem solving program}, in:
  \bibinfo{booktitle}{IFIP congress}, \bibinfo{organization}{Pittsburgh, PA}.
  p.~\bibinfo{pages}{64}.
\bibitem[{Nickel et~al.(2015)Nickel, Murphy, Tresp and
  Gabrilovich}]{nickel2015review}
\bibinfo{author}{Nickel, M.}, \bibinfo{author}{Murphy, K.},
  \bibinfo{author}{Tresp, V.}, \bibinfo{author}{Gabrilovich, E.},
  \bibinfo{year}{2015}.
\newblock \bibinfo{title}{A review of relational machine learning for knowledge
  graphs}.
\newblock \bibinfo{journal}{Proceedings of the IEEE} \bibinfo{volume}{104},
  \bibinfo{pages}{11--33}.
\bibitem[{Paszke et~al.(2019)Paszke, Gross, Massa, Lerer, Bradbury, Chanan,
  Killeen, Lin, Gimelshein, Antiga et~al.}]{paszke2019pytorch}
\bibinfo{author}{Paszke, A.}, \bibinfo{author}{Gross, S.},
  \bibinfo{author}{Massa, F.}, \bibinfo{author}{Lerer, A.},
  \bibinfo{author}{Bradbury, J.}, \bibinfo{author}{Chanan, G.},
  \bibinfo{author}{Killeen, T.}, \bibinfo{author}{Lin, Z.},
  \bibinfo{author}{Gimelshein, N.}, \bibinfo{author}{Antiga, L.}, et~al.,
  \bibinfo{year}{2019}.
\newblock \bibinfo{title}{Pytorch: An imperative style, high-performance deep
  learning library}.
\newblock \bibinfo{journal}{Advances in neural information processing systems}
  \bibinfo{volume}{32}, \bibinfo{pages}{8026--8037}.
\bibitem[{Qu et~al.(2020)Qu, Chen, Xhonneux, Bengio and Tang}]{qu2020rnnlogic}
\bibinfo{author}{Qu, M.}, \bibinfo{author}{Chen, J.},
  \bibinfo{author}{Xhonneux, L.P.}, \bibinfo{author}{Bengio, Y.},
  \bibinfo{author}{Tang, J.}, \bibinfo{year}{2020}.
\newblock \bibinfo{title}{Rnnlogic: Learning logic rules for reasoning on
  knowledge graphs}.
\newblock \bibinfo{journal}{arXiv preprint arXiv:2010.04029} .
\bibitem[{Qu et~al.(2021)Qu, Chen, Xhonneux, Bengio and Tang}]{rnnlogicgithub}
\bibinfo{author}{Qu, M.}, \bibinfo{author}{Chen, J.},
  \bibinfo{author}{Xhonneux, L.P.}, \bibinfo{author}{Bengio, Y.},
  \bibinfo{author}{Tang, J.}, \bibinfo{year}{2021}.
\newblock \bibinfo{title}{Rnnlogic}.
\newblock
  \bibinfo{howpublished}{\url{https://github.com/DeepGraphLearning/RNNLogic}}.
\bibitem[{Qu and Tang(2019)}]{qu2019probabilistic}
\bibinfo{author}{Qu, M.}, \bibinfo{author}{Tang, J.}, \bibinfo{year}{2019}.
\newblock \bibinfo{title}{Probabilistic logic neural networks for reasoning}.
\newblock \bibinfo{journal}{arXiv preprint arXiv:1906.08495} .
\bibitem[{Richardson and Domingos(2006)}]{richardson2006markov}
\bibinfo{author}{Richardson, M.}, \bibinfo{author}{Domingos, P.},
  \bibinfo{year}{2006}.
\newblock \bibinfo{title}{Markov logic networks}.
\newblock \bibinfo{journal}{Machine learning} \bibinfo{volume}{62},
  \bibinfo{pages}{107--136}.
\bibitem[{Sadeghian et~al.(2019a)Sadeghian, Armandpour, Ding and
  Wang}]{drumgithub}
\bibinfo{author}{Sadeghian, A.}, \bibinfo{author}{Armandpour, M.},
  \bibinfo{author}{Ding, P.}, \bibinfo{author}{Wang, D.Z.},
  \bibinfo{year}{2019}a.
\newblock \bibinfo{title}{Drum}.
\newblock \bibinfo{howpublished}{\url{https://github.com/alisadeghian/DRUM}}.
\bibitem[{Sadeghian et~al.(2019b)Sadeghian, Armandpour, Ding and
  Wang}]{sadeghian2019drum}
\bibinfo{author}{Sadeghian, A.}, \bibinfo{author}{Armandpour, M.},
  \bibinfo{author}{Ding, P.}, \bibinfo{author}{Wang, D.Z.},
  \bibinfo{year}{2019}b.
\newblock \bibinfo{title}{Drum: End-to-end differentiable rule mining on
  knowledge graphs}.
\newblock \bibinfo{journal}{arXiv preprint arXiv:1911.00055} .
\bibitem[{Stokman and de~Vries(1988)}]{stokman1988structuring}
\bibinfo{author}{Stokman, F.N.}, \bibinfo{author}{de~Vries, P.H.},
  \bibinfo{year}{1988}.
\newblock \bibinfo{title}{Structuring knowledge in a graph}, in:
  \bibinfo{booktitle}{Human-computer interaction}.
  \bibinfo{publisher}{Springer}, pp. \bibinfo{pages}{186--206}.
\bibitem[{Sun et~al.(2018)Sun, Deng, Nie and Tang}]{rotategithub}
\bibinfo{author}{Sun, Z.}, \bibinfo{author}{Deng, Z.H.}, \bibinfo{author}{Nie,
  J.Y.}, \bibinfo{author}{Tang, J.}, \bibinfo{year}{2018}.
\newblock \bibinfo{title}{Knowledgegraphembedding}.
\newblock
  \bibinfo{howpublished}{\url{https://github.com/DeepGraphLearning/KnowledgeGraphEmbedding}}.
\bibitem[{Sun et~al.(2019)Sun, Deng, Nie and Tang}]{sun2019rotate}
\bibinfo{author}{Sun, Z.}, \bibinfo{author}{Deng, Z.H.}, \bibinfo{author}{Nie,
  J.Y.}, \bibinfo{author}{Tang, J.}, \bibinfo{year}{2019}.
\newblock \bibinfo{title}{Rotate: Knowledge graph embedding by relational
  rotation in complex space}.
\newblock \bibinfo{journal}{arXiv preprint arXiv:1902.10197} .
\bibitem[{Teru and Hamilton(2020)}]{teru2020inductive}
\bibinfo{author}{Teru, K.K.}, \bibinfo{author}{Hamilton, W.L.},
  \bibinfo{year}{2020}.
\newblock \bibinfo{title}{Inductive relation prediction on knowledge graphs}.
\newblock \bibinfo{journal}{ICML, Virtual} .
\bibitem[{Toutanova and Chen(2015)}]{toutanova2015observed}
\bibinfo{author}{Toutanova, K.}, \bibinfo{author}{Chen, D.},
  \bibinfo{year}{2015}.
\newblock \bibinfo{title}{Observed versus latent features for knowledge base
  and text inference}, in: \bibinfo{booktitle}{Proceedings of the 3rd workshop
  on continuous vector space models and their compositionality}, pp.
  \bibinfo{pages}{57--66}.
\bibitem[{Trouillon et~al.(2016)Trouillon, Welbl, Riedel, Gaussier and
  Bouchard}]{trouillon2016complex}
\bibinfo{author}{Trouillon, T.}, \bibinfo{author}{Welbl, J.},
  \bibinfo{author}{Riedel, S.}, \bibinfo{author}{Gaussier, {\'E}.},
  \bibinfo{author}{Bouchard, G.}, \bibinfo{year}{2016}.
\newblock \bibinfo{title}{Complex embeddings for simple link prediction}, in:
  \bibinfo{booktitle}{International conference on machine learning},
  \bibinfo{organization}{PMLR}. pp. \bibinfo{pages}{2071--2080}.
\bibitem[{Wang et~al.(2019a)Wang, Dou, Wu, {de Silva} and
  Jin}]{wangLogicRulesPowered2019}
\bibinfo{author}{Wang, P.}, \bibinfo{author}{Dou, D.}, \bibinfo{author}{Wu,
  F.}, \bibinfo{author}{{de Silva}, N.}, \bibinfo{author}{Jin, L.},
  \bibinfo{year}{2019}a.
\newblock \bibinfo{title}{Logic {{Rules powered knowledge graph embedding}}}.
\newblock \bibinfo{journal}{arXiv:1903.03772 [cs]}
  \href{http://arxiv.org/abs/1903.03772}{\tt arXiv:1903.03772}.
\bibitem[{Wang et~al.(2019b)Wang, Stepanova, Domokos and
  Kolter}]{wang2019differentiable}
\bibinfo{author}{Wang, P.W.}, \bibinfo{author}{Stepanova, D.},
  \bibinfo{author}{Domokos, C.}, \bibinfo{author}{Kolter, J.Z.},
  \bibinfo{year}{2019}b.
\newblock \bibinfo{title}{Differentiable learning of numerical rules in
  knowledge graphs}, in: \bibinfo{booktitle}{International Conference on
  Learning Representations}.
\bibitem[{Wang et~al.(2013)Wang, Mazaitis and Cohen}]{wang2013programming}
\bibinfo{author}{Wang, W.Y.}, \bibinfo{author}{Mazaitis, K.},
  \bibinfo{author}{Cohen, W.W.}, \bibinfo{year}{2013}.
\newblock \bibinfo{title}{Programming with personalized pagerank: a locally
  groundable first-order probabilistic logic}, in:
  \bibinfo{booktitle}{Proceedings of the 22nd ACM international conference on
  Information \& Knowledge Management}, pp. \bibinfo{pages}{2129--2138}.
\bibitem[{Wang et~al.(2019c)Wang, He, Cao, Liu and Chua}]{wang2019kgat}
\bibinfo{author}{Wang, X.}, \bibinfo{author}{He, X.}, \bibinfo{author}{Cao,
  Y.}, \bibinfo{author}{Liu, M.}, \bibinfo{author}{Chua, T.S.},
  \bibinfo{year}{2019}c.
\newblock \bibinfo{title}{Kgat: Knowledge graph attention network for
  recommendation}, in: \bibinfo{booktitle}{Proceedings of the 25th ACM SIGKDD
  International Conference on Knowledge Discovery \& Data Mining}, pp.
  \bibinfo{pages}{950--958}.
\bibitem[{Wei et~al.(2021)Wei, Li, Xin, Wang and Wang}]{mplrgithub}
\bibinfo{author}{Wei, Y.}, \bibinfo{author}{Li, H.}, \bibinfo{author}{Xin, G.},
  \bibinfo{author}{Wang, Y.}, \bibinfo{author}{Wang, B.}, \bibinfo{year}{2021}.
\newblock \bibinfo{title}{Mplr}.
\newblock \bibinfo{howpublished}{\url{https://github.com/lirt1231/MPLR}}.
\bibitem[{Xian et~al.(2019)Xian, Fu, Muthukrishnan, De~Melo and
  Zhang}]{xian2019reinforcement}
\bibinfo{author}{Xian, Y.}, \bibinfo{author}{Fu, Z.},
  \bibinfo{author}{Muthukrishnan, S.}, \bibinfo{author}{De~Melo, G.},
  \bibinfo{author}{Zhang, Y.}, \bibinfo{year}{2019}.
\newblock \bibinfo{title}{Reinforcement knowledge graph reasoning for
  explainable recommendation}, in: \bibinfo{booktitle}{Proceedings of the 42nd
  international ACM SIGIR conference on research and development in information
  retrieval}, pp. \bibinfo{pages}{285--294}.
\bibitem[{Yang et~al.(2014)Yang, Yih, He, Gao and Deng}]{yang2014embedding}
\bibinfo{author}{Yang, B.}, \bibinfo{author}{Yih, W.t.}, \bibinfo{author}{He,
  X.}, \bibinfo{author}{Gao, J.}, \bibinfo{author}{Deng, L.},
  \bibinfo{year}{2014}.
\newblock \bibinfo{title}{Embedding entities and relations for learning and
  inference in knowledge bases}.
\newblock \bibinfo{journal}{arXiv preprint arXiv:1412.6575} .
\bibitem[{Yang et~al.(2017a)Yang, Yang and Cohen}]{yang2017differentiable}
\bibinfo{author}{Yang, F.}, \bibinfo{author}{Yang, Z.}, \bibinfo{author}{Cohen,
  W.W.}, \bibinfo{year}{2017}a.
\newblock \bibinfo{title}{Differentiable learning of logical rules for
  knowledge base reasoning}.
\newblock \bibinfo{journal}{arXiv preprint arXiv:1702.08367} .
\bibitem[{Yang et~al.(2017b)Yang, Yang and Cohen}]{neurallpgithub}
\bibinfo{author}{Yang, F.}, \bibinfo{author}{Yang, Z.}, \bibinfo{author}{Cohen,
  W.W.}, \bibinfo{year}{2017}b.
\newblock \bibinfo{title}{Neural-lp}.
\newblock
  \bibinfo{howpublished}{\url{https://github.com/fanyangxyz/Neural-LP}}.
\bibitem[{Yang and Song(2019)}]{yang2019learn}
\bibinfo{author}{Yang, Y.}, \bibinfo{author}{Song, L.}, \bibinfo{year}{2019}.
\newblock \bibinfo{title}{Learn to explain efficiently via neural logic
  inductive learning}.
\newblock \bibinfo{journal}{arXiv preprint arXiv:1910.02481} .
\bibitem[{Zhang et~al.(2019)Zhang, Paudel, Wang, Chen, Zhu, Zhang, Bernstein
  and Chen}]{zhang2019iteratively}
\bibinfo{author}{Zhang, W.}, \bibinfo{author}{Paudel, B.},
  \bibinfo{author}{Wang, L.}, \bibinfo{author}{Chen, J.}, \bibinfo{author}{Zhu,
  H.}, \bibinfo{author}{Zhang, W.}, \bibinfo{author}{Bernstein, A.},
  \bibinfo{author}{Chen, H.}, \bibinfo{year}{2019}.
\newblock \bibinfo{title}{Iteratively learning embeddings and rules for
  knowledge graph reasoning}, in: \bibinfo{booktitle}{The World Wide Web
  Conference}, pp. \bibinfo{pages}{2366--2377}.
\bibitem[{Zhang et~al.(2020)Zhang, Chen, Yang, Ramamurthy, Li, Qi and
  Song}]{zhang2020efficient}
\bibinfo{author}{Zhang, Y.}, \bibinfo{author}{Chen, X.}, \bibinfo{author}{Yang,
  Y.}, \bibinfo{author}{Ramamurthy, A.}, \bibinfo{author}{Li, B.},
  \bibinfo{author}{Qi, Y.}, \bibinfo{author}{Song, L.}, \bibinfo{year}{2020}.
\newblock \bibinfo{title}{Efficient probabilistic logic reasoning with graph
  neural networks}.
\newblock \bibinfo{journal}{arXiv preprint arXiv:2001.11850} .

\end{thebibliography}

\end{document}